%% file: icml2023_camera_ready.tex
\documentclass{article}
\input{preamble.tex}

\input{preamble_icml.tex}
\input{math}
\input{acronyms}

\begin{document}

\icmltitlerunning{Free-Form Variational Inference for Gaussian Process State-Space Models}
\twocolumn[
\icmltitle{Free-Form Variational Inference for Gaussian Process State-Space Models}



\icmlsetsymbol{equal}{*}

\begin{icmlauthorlist}
\icmlauthor{Xuhui Fan}{nc}
\icmlauthor{Edwin V. Bonilla}{data}
\icmlauthor{Terence J. O’Kane}{env}
\icmlauthor{Scott A. Sisson}{unsw}
\end{icmlauthorlist}

\icmlaffiliation{nc}{University of Newcastle, Australia}
\icmlaffiliation{data}{CSIRO's Data61, Australia}
\icmlaffiliation{env}{CSIRO's Environment, Australia}
\icmlaffiliation{unsw}{University of New South Wales, Australia}

\icmlcorrespondingauthor{Xuhui Fan}{\mbox{xhfan.ml@gmail.com}}

\icmlkeywords{Machine Learning, ICML}

\vskip 0.3in
]



\printAffiliationsAndNotice{}  % leave blank if no need to mention equal contribution

\begin{abstract}
Gaussian process state-space models (GPSSMs) provide a principled and flexible approach to modeling the dynamics of a latent state, which is observed at discrete-time points via a likelihood model. 
However, inference in GPSSMs is computationally and statistically challenging due to the large number of latent variables in the model and the strong  temporal dependencies between them. 
In this paper, we propose a new  method for inference in Bayesian GPSSMs, which overcomes the drawbacks of previous approaches, namely over-simplified assumptions, and high computational requirements.  
Our method is based on free-form variational inference via stochastic gradient Hamiltonian Monte Carlo within the inducing-variable formalism. Furthermore, by exploiting our proposed variational distribution, we provide a collapsed extension of our method where the inducing variables are marginalized analytically. We also showcase results when combining our framework with particle MCMC methods. We show that, on six real-world datasets, our approach can learn transition dynamics and latent states more accurately than competing methods. 
\end{abstract}

\section{Introduction}
\label{sec:intro}
% TODO:
% Cite paper on understanding the KF (statistics) 

% 1. WHAT STATE-SPACE MODELS ARE AND WHY THEY ARE IMPORTANT 
\Acrlongpl{SSM} \citep[\acrshortpl{SSM};][Ch.~29]{pml2Book} characterize the underlying dynamics of a latent state given a set of observations via a transition %(or evolution) 
function and an observation model.  
As a  modeling framework, they provide a general approach for understanding time-series data \citep{state-space-models} and for data assimilation problems \citep{enkf-understanding-2016}. 
Applications of \acrshortpl{SSM} abound and span diverse areas such as econometrics \citep{tsay2005analysis}, %meteorology \citep{hernandez1991state}, 
control engineering \citep{ogata2010modern} and neuroscience \citep{brown1998statistical}. 

% 2. WE FOCUS ON GP-BASED TRANSITION FUNCTIONS (GPSSMS)
In this paper, we  focus on Bayesian \acrshortpl{SSM}, where the transition function describing the dynamics of the system is given a prior distribution.
 A paradigmatic example of Bayesian \acrshortpl{SSM} are \acrlongpl{GPSSM} \citep[\acrshortpl{GPSSM};][]{frigola2015thesis},  where this prior distribution  is  a \acrlong{GP} \citep[\acrshort{GP};][]{williams2006gaussian}. 
Due to their Bayesian non-parametric nature, \acrshortpl{GPSSM} represent a principled and flexible approach to Bayesian \acrshortpl{SSM}. 
 
% 3. INFERENCE IN GPSSMs IS INCREDIBLY HARD, FROM THE STATISTICAL AND COMPUTATIONAL PERSPECTIVES 
However, the flexibility of \glspl{GP} adds significant computational and statistical challenges to the already difficult problem of inference in Bayesian \acrshortpl{SSM}. Indeed, even for non-Bayesian \acrshortpl{SSM}, standard problems such as filtering, smoothing, and prediction are, in general, analytically intractable\footnote{With the notable exception of % linear state transition and observation models with additive Gaussian noise, for which the solution to the optimal filtering problem is given by the  Kalman filter \citep{Kalman1960a}.}
the linear-Gaussian case where the optimal solution is given by the 
Kalman filter \citep{Kalman1960a}.}. 
Having a \acrshort{GP} prior over the transition function in \acrshortpl{SSM} increases the number of latent variables significantly; incorporates strong (and potentially long-term) dependencies across states; and introduces a cubic time complexity as a function of the number of observations. 

Within the \acrshort{GP} community, significant advances have been made addressing 
the computational issues in \acrshort{GP} regression and classification problems, most notably using inducing-variable approximations  \cite{titsias2009variational,hensman-mcmc-2015,pmlr-v130-rossi21a} but also random-feature expansions \cite{pmlr-v70-cutajar17a,marmin2022deep} and, more recent innovative approaches such as the Vecchia approximation \cite{sauer-et-al-2022}. Although these approximations are applicable to \acrshort{GP}-based dynamic models, the challenges above remain prevalent within the context of \acrshortpl{GPSSM}. 

% 4. PREVIOUS APPROACHES
% based on the inducing variables 
% previous approaches: oversimplified assumptions (decoupled, or assumptions on the posterior) or high-computational requirements 
Nevertheless, previous approaches have developed insightful and practical algorithms for inference in \acrshortpl{GPSSM}, mainly based on inducing-variable  approximations, which is also our main underpinning methodology for scalable \acrshortpl{GP}. 
In outlining the most relevant approaches, our main object of interest is the approximate joint posterior over state trajectories $\mbx_{0:T}$ and inducing variables $\mbu$, $q(\mbx_{0:T}, \mbu)$, where $T+1$ is the length of the trajectory. 
We consider two main aspects of this joint distribution: (i) whether the dependencies between state trajectories and inducing variables are captured and (ii) whether their corresponding distributions are unconstrained, i.e., not restricted to a sub-optimal parametric form. 
The seminal \gls{VGPSSM} proposed by \citet{vi-gpssm} as well as the subsequent \gls{IGPSSM} of \citet{eleftheriadis2017identification}  use mean-field approaches, therefore, ignoring the posterior dependencies between state trajectories and inducing variables. 
The more recent  methods, namely the \gls{PRSSM} of \citet{PRSSM} and the \gls{VCDT} of \citet{overcome-mean-field-gp} introduce couplings across state trajectories and inducing variables. However, their posteriors are constrained to be Gaussians. 
Thus, these previous works have either assumed independence between state trajectories and inducing variables or imposed strong parametric constraints in their corresponding posteriors or both.  

As  shown  by \citet{overcome-mean-field-gp}, a mean-field posterior can yield poor practical performance. 
Similarly, a Gaussian assumption on the state posterior or the inducing variable posterior is also very strong and, by definition, will not generally capture the true posterior even in the limit of infinite computation.  
To address these issues,  we propose a free-form variational inference approach to posterior estimation in \acrshortpl{GPSSM} that models the full joint distribution over states and inducing variables, $q(\mbx_{0:T}, \mbu)$, without any mean-field or parametric assumptions.  
We refer to our method as \gls{FFVD}
% \footnote{Pronounced as ``effvid''.} 
and summarize the major differences between its posterior assumptions and those of previous approaches in \cref{tab:main-method-comparison}. Below we describe our contributions in more detail. 

\textbf{(i) Flexible posterior:} We develop \gls{FFVD}, an inference algorithm for \acrshortpl{GPSSM} based on \acrlong{SGHMC} \citep[\acrshort{SGHMC};][]{chen2014stochastic,havasi2018inference}, which represents the posterior over states and inducing variables using samples. 
\gls{FFVD} lifts the limitations of previous variational approaches to \acrshortpl{GPSSM}, which have ignored couplings in this posterior %between the inducing variables and the state trajectories
or have assumed a constrained parametric form. % for this posterior. 
More precisely, \gls{FFVD} %naturally 
captures the posterior correlations between states and inducing variables; does not constrain this posterior to any parametric form, and, is scalable to a large number of observations. 

\textbf{(ii) Collapsed inference that accelerates convergence:} we show that (i) our formulation allows us to collapse the inducing variables $\mbu$, (ii) sample from the lower-dimensional marginal $q(\mbx_{0:T})$ %using \gls{SGHMC} 
and, at the end of the sampling procedure, (iii) obtain samples from the conditional $q(\mbu \g \mbx_{0:T})$, for which we derive a closed-form expression.  We show that collapsing accelerates convergence significantly. %, compared to the uncollapsed version. 

\textbf{(iii) Extensions with \gls{PMCMC}}: we further investigate whether more elaborate inference algorithms, such as \gls{PMCMC} \citep{particlemcmc}, that account for the sequential nature of the problem can provide more accurate posteriors.

\textbf{(iv) State-of-the-art performance:} we showcase the properties and benefits of our approach compared to previous methods such as \gls{VGPSSM}, \gls{PRSSM} and \gls{VCDT} in a synthetic example and six system identification benchmarks. Overall, our method provides state-of-the-art performance when evaluated on these problems, while having comparable  computational requirements to previous approaches. 
Our code and supplementary material can be found at  \url{https://github.com/xuhuifan/FFVD}.

\begin{table*}[h]
    \centering
    \caption{Comparison across  methods in terms of their assumptions on the variational distribution. 
    The  rows refer to whether the variational posterior captures the dependencies between the state trajectories and the inducing variables (coupled $q({\mbx}_{0:T}, {\mbu})$); 
    the distribution over states is unconstrained (unconstrained $q({\mbx}_{0:T} \g{\mbu})$ or $q({\mbx}_{0:T})$), i.e., not restricted  to a sub-optimal parametric form; and whether the distribution over the inducing variables is also unconstrained (unconstrained $q({\mbu})$). 
    % For the complexities, $T_{\text{train}}$ and $T_{\text{test}}$ are the lengths of the training and testing sequences, respectively; $M$ is the number of inducing variables, and $S$ is the number of posterior samples.
    Our method is referred to as \acrshort{FFVD}.
    %The different methods appear on the columns, with ours referred to as \gls{FFVD}. 
    }
    \label{tab:main-method-comparison}
    \begin{tabular}{l c c c c c c}
    \toprule
& \acrshort{VGPSSM} & \acrshort{IGPSSM} & \acrshort{PRSSM} & \acrshort{VCDT} &   \acrshort{FFVD}\\
\midrule
Coupled $q({\mbx}_{0:T}, {\mbu})$ & \xmark & \xmark & \cmark & \cmark   & \cmark \\
Unconstrained $q({\mbx}_{0:T} \g {\mbu})$ or $q({\mbx}_{0:T}$) & \cmark &  \xmark & \xmark & \xmark & \cmark \\
Unconstrained $q(\mbu)$ & \cmark & \xmark & \xmark &  \xmark &  \cmark \\
% \midrule 
% Training complexity & $\mathcal{O}(M^2T_{\text{train}})$ & $\mathcal{O}(M^2T_{\text{train}})$ & $\mathcal{O}(M^2T_{\text{train}})$ & $\mathcal{O}(M^2T_{\text{train}})$  &  \\
% Prediction complexity & $\mathcal{O}(M^2T_{\text{test}})$  & $\mathcal{O}(T^2_{\text{train}} S \cdot T_{\text{test}})$  & & $\mathcal{O}(M^2T_{\text{test}})$  &   \\
% Memory complexity     &  &    &   &  &   \\
\bottomrule
    \end{tabular}
\end{table*}
\section{Inference in Gaussian Process Models}
\label{sec:GPs}
\glsresetall % Resetting all acronyms so we remind the reader what things stan for 
\glspl{GP} are priors over functions where every subset of function values follow a Gaussian distribution. 
We use $f(\mbx) \sim \GP\left(\mean(\mbx), \kernel(\mbx,\mbx^\prime; \gphyper)\right)$ to denote that $f$ is distributed according to a \gls{GP} with mean function $\mean(\cdot)$ and covariance function $\kernel(\cdot,\cdot; \gphyper)$, where $\gphyper$ are referred to as the \gls{GP} hyper-parameters\footnote{% note here that the mean function can also have additional parameters but, in this work,
We have simply assumed identity mean functions.}.
By definition, a \gls{GP} prior over functions implies a finite prior over $T$ function values $\mbf = [f(\mbx_1), \ldots, f(\mbx_T)]^\top$, i.e., $\mbf \sim \Normal(\mbf; \mbm, \mbK)$, where $\mbm$ and $\mbK$ are obtained by evaluating the mean function and covariance function at all the inputs $\mbX := \{\mbx_1, \ldots, \mbx_T\}$. 

In supervised learning settings, we are given input-output observations $\{\mbx_t, y_t\}_{t=1}^T$ and a conditional likelihood model $p(\mby \g \mbf)$. Inference involves estimating the posterior distribution $p(\mbf \g \mby, \mbX)$ and the hyper-parameters $\gphyper$ from data. Consequently, we can use these to estimate the posterior predictive distribution at a new point $\mbx_\star$, i.e., $p(f(\mbx_\star) \g \mbX, \mby, \mbx_\star)$.  Notoriously, these tasks have  cubic time complexity as a function of the number of training observations, i.e., $\cO(T^3)$, arising from algebraic operations involving the computation of the inverse covariance and its log determinant. This motivates the need for sparse approximations.
\subsection{Sparse \gls{GP} Approximations via Inducing Variables}
%
% Various approximations have been proposed 
To deal with the cubic computational complexity of inference in \gls{GP} models, % In this paper, 
we focus on inducing-variable approximations based on variational inference, as originally proposed by \citet{titsias2009variational} and made scalable to very large datasets by \citet{hensman-uai-2013}. 
The main idea of these approximations is to augment the space of function values with a set of $M$ inducing variables $\mbu:= \{u_i\}$ and their corresponding inducing inputs $\mbZ := \{\mbz_i\}$. 
Thus, inference involves estimating the posterior $q(\mbu, \mbf) \approx p(\mbu, \mbf \g \mbX, \mby)$ and the \gls{GP} hyper-parameters $\gphyper$ via variational inference. 
Under the assumption that $q(\mbu, \mbf):=q(\mbu) p(\mbf \g \mbu)$ where  $p(\mbf \g \mbu)$ is the conditional prior, the variational objective, the so-called \gls{ELBO}, decomposes over the observations, and inference can be carried out with time complexity of $\cO(M^3)$, providing significant advantages when $M \ll T$. 

\section{Gaussian Process State-Space Models}
\label{sec:gpssm}
Let us assume we  are given a time series of $T$ multi-dimensional observations $\mby_{1:T}$ and denote their corresponding latent states with $\mbx_{0:T}$, where $\mbx_0$ is the initial state.  Here we denote the time series  $\mbx_{t_1:t_2} := \{ \mbx_{t_1}, \ldots, \mbx_{t_2} \}$, and similarly for $\mby_{1:T}$. 
In general, $\mbx_t \in \bbR^{d_x}$ and $\mby_t \in \bbR^{d_y}$. 
\Glspl{GPSSM} formulate a discrete-time \gls{SSM} where the transition dynamics is given by a \gls{GP}.  The full generative process is:
\begin{align}
    \mbx_0  & \sim p(\mbx_0), \text{ }
    f(\mbx)  \sim \GP(\mean(\mbx),\kernel(\mbx, \mbx^\prime;\gphyper)), \\
    \mbf_t  &:= f(\mbx_{t-1}),  \quad 
    \label{eq:transition-model}
    \mbx_t \g \mbf_t  \sim \Normal(\mbx_t;\mbf_t,\mbQ), \\
    \mby_t \g \mbx_t & \sim p(\mby_t\g\mbx_t,\mbphi),
\end{align}
where $\mbQ$ is the transition process covariance, 
% measuring the dispersion of the current state $\mbx_t$ with respect to the function value $\mbf_t$ 
and $\mbphi$ is the vector of parameters of the conditional likelihood $p(\mby_t \g \mbx_t,\mbphi)$. 
%
%\subsection{Conditional Likelihood Model}
%\label{sec:cond-likelihood-gaussian}
Although our framework does not make any parametric assumptions about this conditional likelihood, in our experiments in \cref{sec:experiments}, we adopt the same setting as in previous works \citep{overcome-mean-field-gp,PRSSM} and set $p(\mby_t\g\mbx_t,\mbphi) := \Normal(\mby_t;\mbC\mbx_t+\mbd,\mbR)$, with $\mbphi=\{\mbC,\mbd,\mbR\}$, where $\mbC,\mbd$ are the weights and bias of the linear transformation on $\mbx_t$ and $\mbR$ is the observation covariance. 
% \Cref{fig:generative_process_GPSSM} shows the graphical model for this full \gls{GPSSM}. 
%
\subsection{Joint Distribution}
\label{sec:full-gp}
As shown by \citet{frigola2015thesis}, to sample $\mbf_t$, instead of conditioning on an infinite-dimensional function, we can condition only on the transitions seen up to (but not including) time $t$, i.e., $\{(\mbx_{i-1}, \mbf_i)\}_{i=1}^{t-1}$. We can then write the joint distribution over latent variables and observations as:
\begin{multline}
    \label{eq:joint-full}
    p(\mby_{1:T}, \mbx_{0:T}, \mbf_{1:T}) =\\ 
    p(\mbx_0) 
    \prod_{t=1}^{T} p(\mbf_t \g \mbf_{1:t-1}, \mbx_{0:t-1}) p(\mbx_t \g \mbf_t) p(\mby_t \g \mbx_t) ,
\end{multline}
where $p(\mbx_0), p(\mbx_t \g \mbf_t), p(\mby_t \g \mbx_t)$ are defined as above and, for the edge case of $t=1$, we have: $p(\mbf_1 \g \mbx_0) = \Normal(\mbf_1; \mean(\mbx_0), \kernel(\mbx_0, \mbx_0; \gphyper))$. 
Furthermore, we recognize each conditional distribution $p(\mbf_t \g \mbf_{1:t-1}, \mbx_{0:t-1})$ in \cref{eq:joint-full}  as the \gls{GP} prediction at a single point $\mbx_{t-1}$ using noiseless outputs $\mbf_{1:t-1}$ and inputs $\mbx_{0:t-2}$, 
\begin{align}
     p(\mbf_t \g \mbf_{1:t-1}, \mbx_{0:t-1}) &=
     p(\mbf_t \g \mbx_{t-1}, \mbf_{1:t-1}, \mbx_{0:t-2}) \\
   & =  \Normal(\mbf_t; \mbmu_f , \mbSigma_f),
\end{align}
with conditional mean and covariance given by 
\begin{align}
   \mbmu_f &=  \mx{t-1} + \Kxy{t-1}{0:t-2} \Kxx{0:t-2}^{-1} (\mbf_{1:t-1} - \mx{0:T-2}) \\
   \mbSigma_f &=  \Kxx{t-1} - \Kxy{t-1}{0:t-2} \Kxx{0:t-2}^{-1} \Kxy{0:t-2}{t-1}), 
    \label{eq:full-cond}
\end{align}
where the subscript notation indicates the mean vectors and covariance matrices obtained from evaluating the mean function and covariance function, respectively,  at the corresponding ranges,  $\mx{t_1:t_2} := \mean(\mbx_{t_1:t_2})$, $\mx{t} := \mx{t:t}$, $\Kxy{t_1:t_2}{t_3:t_4} := \kernel(\mbx_{t_1:t_2}, \mbx_{t_3:t_4}; \gphyper)$, $\Kxx{t_1:t_2} := \Kxy{t_1:t_2}{t_1:t_2}$ and $\Kxx{t} := \Kxx{t:t}$.  
\subsection{Multidimensional Latent States \& Control Inputs}
In the case of multidimensional latent states, i.e., $d_x > 1$, we assume independent \glspl{GP} on each dimension, each with its own mean function and covariance function. 
Because of this independence assumption, each \gls{GP} only has to condition on its own function evaluations. % based on its own mean function, covariance function, and inducing points. 
Therefore, for simplicity in the notation, we do not index the means and covariances with respect to their dimension $d$ and consider the underlying  \glspl{GP}, their means, and covariance functions as multi-dimensional.    
% \eb{clarify a little bit more here based on my handwritten notes}
%
% \subsubsection{Control Inputs} 
Furthermore, as we shall see in \cref{sec:experiments}, our experiments consider additional control inputs ($\mba_t$) that affect the transitions in a Markovian way. This is easy to incorporate in our framework by augmenting our input space and, therefore, indexing the \glspl{GP} in the higher-dimensional space given by the concatenation $[\mbx^\top,\mba^\top]^\top$. However, since the control inputs are fixed and deterministic, we do not need to include them as part of out inference method and  do not make them explicit in our subsequent mathematical development.  
%------------
%
% \cite{frigola2015thesis} shows that it is feasible to condition only on the transitions up to (but not including) time $t$, i.e., $\{(\mbx_{i-1},\mbf_i)\}_{i=1}^t$, to sample $\mbf_t$, rather than conditioning on an infinite-dimensional function $f(\cdot)$. That is, we may define $p(\mbf_{t}\g\mbf_{1:(t-1)},\mbx_{0:t-1})$ as a conditional distribution to predict the function value $\mbf_t$ at a single point $\mbx_{t-1}$ using noiseless outputs $\mbf_{1:t-1}$ and inputs $\mbx_{0:t-1}$, which is:
% \begin{align}
%     &p(\mbf_{t}\g\mbf_{1:(t-1)},\mbx_{0:t-1}):\nonumber \\
%     =&\Normal\left(
%     \mbf_t;\mx{t-1}+\Kxy{t-1}{0:t-2}\Kxx{0:t-2}^{-1}(\mbf_{1:t-1}-\mx{0:t-2}),\right.\nonumber \\
%     &\left.\qquad\Kxx{t-1}-\Kxy{t-1}{0:t-2}\Kxx{0:t-2}^{-1}\Kxy{0:t-2}{t-1}\right)
% \end{align}
% in which $\mx{t_1:t_2}\defeq\mean(\mbx_{t1:t2})$, $\mx{t}\defeq\mx{t:t}$, $\Kxy{t_1:t_2}{t_3:t_4}\defeq\kernel(\mbx_{t_1:t_2},\mbx_{t_3:t_4};\gphyper)$, $\Kxx{t_1:t_2}\defeq\Kxy{t_1:t_2}{t_1:t_2}$, and $\Kxx{t}\defeq\Kxx{t:t}$

% The joint distribution can thus be further written as:
% \begin{align}
%     p(\mby_{1:T},\mbx_{0:T}, \mbf_{1:T})=p(\mbx_0)\prod_{t=1}^T[&p(\mbf_{t}\g\mbf_{1:(t-1)},\mbx_{0:t-1})\nonumber\\ 
%     &\cdot p(\mbx_t\g\mbf_t)p(\mby_t\g\mbx_t)]\nonumber
% \end{align}
% For the edge case of $t=1$, we have $p(\mbf_t\g\mbx_0)=\Normal(\mbf_1;\mean(\mbx_0),\kernel(\mbx_0,\mbx_0;\gphyper))$. 
%
\subsection{Sparse \gls{GPSSM}}
\label{sec:sparse-gp}
In order to make the \gls{GPSSM} inherently scalable, similarly to the standard supervised regression setting with \glspl{GP} described in \cref{sec:GPs}, we augment the full \gls{GPSSM} model with $M$ inducing variables and corresponding inducing inputs $\mbu:=\{ u_i \}_{i=1}^M$ and $\mbZ := \{ \mbz_i \}_{i=1}^M$.
This gives rise to  the sparse \gls{GPSSM} model:
\begin{multline}
    p(\mby_{1:T}, \mbx_{0:T}, \mbf_{1:T}, \mbu \g \mbZ) = 
    p(\mbu \g \mbZ) 
    p(\mbx_0) \\
 \prod_{t=1}^{T} p(\mbf_t \g \mbf_{1:t-1}, \mbx_{0:t-1}, \mbu, \mbZ) p(\mbx_t \g \mbf_t) p(\mby_t \g \mbx_t),  
    \label{eq:joint-sparse}    
\end{multline}
where the prior over the inducing variables is determined by the \gls{GP} prior, i.e.,   $p(\mbu \g \mbZ) = \Normal(\mz, \Kzz)$ 
with $\mz \defeq m_f(\mbZ)$ and $\Kzz \defeq \kernel(\mbZ, \mbZ; \gphyper)$. 
Besides having a prior over the inducing variables $\mbu$, the main difference with our previous full-model formulation of \cref{sec:full-gp} is that the conditional distributions over $\mbf_t$ have been augmented with inducing variables $\mbu$ and corresponding inducing inputs $\mbZ$. 
Each conditional, $p(\mbf_t \g \mbf_{1:t-1}, \mbx_{0:t-1}, \mbu, \mbZ) = p(\mbf_t \g \mbx_{t-1}, \mbf_{1:t-1}, \mbu, \mbx_{0:t-2},  \mbZ)$, is the predictive (\gls{GP} regression) distribution of $\mbf_t$ at test input $\mbx_{t-1}$ when observing $(\mbf_{1:t-1}, \mbu)$ at their respective locations $(\mbx_{0:t-2}, \mbZ)$. 
% -------------
% \noindent\textbf{Sparse GPSSMs} We may augment GPSSMs with inducing variables $\mbu\defeq\{u_i\}_{i=1}^M$ and their corresponding inducing inputs $\mbz\defeq\{z_i\}_{i=1}^M$:
% \begin{align} \label{eq:prior}
% p(\mbu\g\mbz)=&\Normal(\mx{\mbz}, \Kxx{\mbz})\nonumber \\
%     p(\mby_{1:T},\mbx_{0:T}, \mbf_{1:T}, \mbu\g\mbz)=&\prod_{t=1}^T[p(\mbf_{t}\g\mbf_{1:(t-1)},\mbx_{0:t-1},\mbu,\mbz)\nonumber\\ &p(\mbx_t\g\mbf_t)p(\mby_t\g\mbx_t)]p(\mbu\g\mbz)p(\mbx_0)
% \end{align}
% where $\mx{\mbz}=\mean(\mbz), \Kxx{\mbz}\defeq\kernel(\mbz,\mbz)$. Besides having a prior over the inducing variables $\mbu$, the main difference between sparse GPSSMs and GPSSMs is that the conditional distribution over $\mbf_t$ has been augmented with inducing variables $\mbu$ and inducing inputs $\mbz$. Each conditional, $p(\mbf_{t}\g\mbf_{1:(t-1)},\mbx_{0:t-1},\mbu,\mbz)$ is the predictive distribution of $\mbf_t$ at test point $\mbx_{t-1}$ when observing $(\mbf_{1:t-1};\mbu)$ the their retrospective locations $(\mbx_{0:t-2};\mbz)$.
%
\section{Free-Form Variational Inference} 
\label{sec:main-methods}
In this section, we develop a posterior estimation method using \gls{VI}. We show that under a specific form of the approximate posterior, we can estimate the joint posterior over $\{\mbx_{0:T},\mbu\}$ in free-form, i.e., optimally, without making any assumptions such as independence between state trajectories and inducing variables, typical of mean-field approaches \citep{vi-gpssm,eleftheriadis2017identification}, or constraining the form of the posterior to sub-optimal parametric forms \citep{overcome-mean-field-gp,PRSSM}. 
%
% ~\cite{vi-gpssm}, prior~\cite{PRSSM}, Gaussian~\cite{eleftheriadis2017identification,overcome-mean-field-gp}, about the structure. 
%
\subsection{Variational Family and Evidence Lower Bound}
\label{sec:var-family}
\Acrlong{VI} is underpinned by the maximization of the \acrfull{ELBO}, which is equivalent to minimizing the \gls{KL} divergence between the approximate posterior and the true posterior. This objective is  given by
\begin{multline} 
\label{eq:elbo}
 %    \cL_{\text{ELBO}} :=
 % \E_{q(\mbx_{0:T}, \mbu, \mbf_{1:T})}\left[\log \frac{p(\mby_{1:T},\mbx_{0:T},\mbu,\mbf_{1:T}\g\mbZ)}{q(\mbx_{0:T},\mbu,\mbf_{1:T})}\right],
     \cL_{\text{ELBO}}(q) :=
 \E_{q(\mbx_{0:T}, \mbu, \mbf_{1:T})}
 \left[
 \log {p(\mby_{1:T},\mbx_{0:T},\mbf_{1:T},\mbu\g\mbZ)} \right.\\
 \left.
 - \log {q(\mbx_{0:T},\mbu,\mbf_{1:T})}
 \right],
\end{multline}
where   $p(\mby_{1:T}, \mbx_{0:T}, \mbf_{1:T}, \mbu \g \mbZ)$ is the joint distribution in \cref{eq:joint-sparse} and $q(\mbx_{0:T},\mbu,\mbf_{1:T})$ is our proposed approximate joint posterior, which we define as
\begin{align}
q(\mbf_{1:T}\g\mbx_{0:T},\mbu) 
& \defeq 
\prod_{t=1}^Tp(\mbf_{t}\g\mbf_{1:t-1},\mbx_{0:t-1},\mbu,\mbZ), \label{eq:proposal_f}\\    
q(\mbx_{0:T},\mbu,\mbf_{1:T}) 
& := 
q(\mbx_{0:T},\mbu)q(\mbf_{1:T}\g\mbx_{0:T},\mbu) \label{eq:proposal_ufx}.
\end{align}
It is easy to show that the \gls{ELBO}, as defined in \cref{eq:elbo}, is a lower bound on the log marginal likelihood, i.e., $\cL_{\text{ELBO}} (q) \leq \log p(\mby\g\mbZ)$. 
We see that our joint variational posterior over state trajectories $\mbx_{0:T}$, inducing variables $\mbu$ and latent function values $\mbf_{1:T}$ in \cref{eq:proposal_ufx}
uses the conditional prior over the latent function values in \cref{eq:proposal_f}. This is, of course, an assumption and limits the flexibility of our approximate posterior. However, all previous scalable variational approaches to \glspl{GPSSM}, such as \acrshort{VGPSSM}, \acrshort{IGPSSM}, \acrshort{PRSSM} and \acrshort{VCDT}, have made the very same assumption.  
Indeed, these types of approximations where the conditional prior is used to define the join variational posterior have become customary and \textit{necessary} to avoid the cubic time complexity of inference in \gls{GP} models. 
See \cref{sec:discussion} for further discussion on the limitations of our approach.
%It is necessary to make the variational framework scalable as a function of the number of observations, which will yield an inference algorithm that alleviates the cubic time complexity on the number of observations $(T)$ and, therefore, scales up to large datasets. 

Nevertheless, our proposed joint variational distribution in \cref{eq:proposal_ufx}, will allow us to derive an \emph{optimal} variational distribution $q(\mbx_{0:T},\mbu)$ in free-form,  without imposing any 
parametric constraints over it and, instead, represent it via samples. 
We will describe this in the next section. 

% as we did not set assumptions on its structure. Assuming the conditional distribution of $\mbf_{1:T}$ is the same as its prior is the main assumption in the variational distribution. This assumption is also used in previous approaches. This conditional distribution would appear in both the numerator and denominator in Eq.~(\ref{eq:elbo}) and can be cancelled out. Since we are using the sparse methods, we mainly use the inducing variables $\mbu$ to make predictions about the function value at new latent states, and the prediction does not involve the function values at existing latent states. 
%
\subsection{Evidence Lower Bound Maximization}
Our first step is to expand the expression for the \gls{ELBO} in \cref{eq:elbo} using our joint model distribution in \cref{eq:joint-sparse} and our proposed variational distribution in \cref{eq:proposal_f,eq:proposal_ufx}.  We first note that our definition of the variational distribution in \cref{eq:proposal_ufx} uses the same conditional prior as in the joint distribution in \cref{eq:joint-sparse}. Therefore, this term cancels out, avoiding the computation of operations on fully-coupled high-dimensional distributions over latent functions. Thus, we have that:
%
% By summarizing Eq.~(\ref{eq:prior})(\ref{eq:elbo})(\ref{eq:proposal_f})(\ref{eq:proposal_ufx}), we can have the Evidence Lower Bound as follows:
\begin{align}
    % &\cL_{\text{ELBO}}\nonumber \\
    % =&\int q(\mbx_{0:T},\mbu)\log\frac{p(\mbx_0)p(\mbu\g\mbZ)\prod_{t=1}^T p(\mby_t\g\mbx_t)}{q(\mbx_{0:T},\mbu)}\diff\mbx_{0:T}\diff\mbu\nonumber \\
    % &  +
    %  \int 
    %  q(\mbx_{0:T}, \mbu) \E_{q(\mbf_{1:T} \g \mbx_{0:T}, \mbu )}
    %  [\log \prod_{t=1}^T p(\mbx_t \g \mbf_t) ]
    %   \diff\mbx_{0:T} \diff\mbu  
    % Other version 
    \nonumber
    & \cL_{\text{ELBO}} (q) =  \int q(\mbx_{0:T},\mbu)
    \Bigg\{ -   \log q(\mbx_{0:T},\mbu)  \\ 
    \nonumber
    & +  \log \Big[
        p(\mbx_0)p(\mbu\g\mbZ)\prod_{t=1}^T p(\mby_t\g\mbx_t) \Big] \\ 
    &  
    +  \E_{q(\mbf_{1:T} \g \mbx_{0:T}, \mbu )}
     \Big[\log \prod_{t=1}^T p(\mbx_t \g \mbf_t) \Big]
     \Bigg\}
      \diff\mbx_{0:T} \diff\mbu .
\end{align}
%Maximizing $\cL_{\text{ELBO}}$ can be done by solving the following Euler-Lagrange Equation:
% \subsubsection{\gls{ELBO} Maximization}
Next we aim to maximize the \gls{ELBO} functional above with respect to $q(\mbx_{0:T-1}, \mbu)$  subject to the constraint 
$\int q(\mbx_{0:T-1}, \mbu) \ \diff\mbx_{0:T-1} \diff\mbu = 1$. We can do this by solving the corresponding Euler-Lagrange equation:
\begin{multline}
    % % &\parderiv{\cL_{\text{ELBO}}}{q(\mbx_{0:T},\mbu)} =
    % \parderiv{}{q(\mbx_{0:T},\mbu)} % \nonumber \\
    % & \left[\int q(\mbx_{0:T},\mbu)\log\frac{p(\mbx_0)p(\mbu\g\mbZ)\prod_t p(\mby_t\g\mbx_t)}{q(\mbx_{0:T},\mbu)}\diff\mbx_{0:T}\diff\mbu\right.\nonumber \\
    % &  \left.+
    %  \int 
    %  q(\mbx_{0:T}, \mbu) \E_{q(\mbf_{1:T} \g \mbx_{0:T}, \mbu )}
    %  [\log \prod_{t} p(\mbx_t \g \mbf_t) ]
    %   \diff\mbx_{0:t} \diff\mbu \right]\nonumber \\
    %   =&0
    % &\parderiv{\cL_{\text{ELBO}}}{q(\mbx_{0:T},\mbu)} =
    \parderiv{}{q(\mbx_{0:T},\mbu)} % \nonumber \\
     \Bigg\{
     - q(\mbx_{0:T},\mbu)\log q(\mbx_{0:T},\mbu) \\
    +  q(\mbx_{0:T},\mbu) 
   \log \Big[ p(\mbx_0)p(\mbu\g\mbZ)\prod_{t=1}^T p(\mby_t\g\mbx_t) \Big] \\
      +
     q(\mbx_{0:T}, \mbu) \E_{q(\mbf_{1:T} \g \mbx_{0:T}, \mbu )}
     \Big[\log \prod_{t=1}^T p(\mbx_t \g \mbf_t) \Big]
       \Bigg\} = 0.
\end{multline}
By doing the corresponding derivatives we obtain
\begin{multline}
    - \log q(\mbx_{0:T},\mbu) - 1 + 
    \log \Big[ p(\mbx_0)p(\mbu\g\mbZ)\prod_{t=1}^T p(\mby_t\g\mbx_t) \Big] \\
     + \E_{q(\mbf_{1:T} \g \mbx_{0:T}, \mbu )}
     \Big[ \log \prod_{t=1}^T p(\mbx_t \g \mbf_t) \Big]=0.
\end{multline}
Now the expectation above can be solved in closed form:
\begin{align}
    & \E_{q(\mbf_{1:T} \g \mbx_{0:T}, \mbu )}
     \Big[\log \prod_{t=1}^T p(\mbx_t \g \mbf_t) \Big] \nonumber\\
     =& \sum_{t=1}^T \E_{p(\mbf_{t} \g \mbx_{t-1}, \mbu, \mbZ )}
     \log p(\mbx_t \g \mbf_t) \nonumber\\
     =&\sum_{t=1}^T \left[\log \Normal(\mbx_t; \mbu_{x_t}, \mbQ)-\frac{1}{2}\trace(\mbQ^{-1}\mbB_{t-1})\right],
\end{align}
where  $p(\mbf_t \g  \mbx_{t-1}, \mbu, \mbZ)$ is the \gls{GP} predictive distribution over the function values $\mbf_{t}$ at locations $ \mbx_{t-1}$ given the  inducing variables $\mbu$ at inducing inputs $\mbZ$, i.e., 
\begin{align}
\label{eq:marg-conditional}
p(\mbf_t \g  \mbx_{t-1}, \mbu, \mbZ) & = \Normal(\mbf_t; \mbmu_{x_t} , \mbB_{t-1} ), \\
\label{eq:muxt}
\mbmu_{x_t} & :=\mx{t-1} + \mbA_{t-1}(\mbu - \mz).
\end{align}
Here we have defined 
\begin{align}
    \label{eq:At}
 \mbA_{t-1} & := \Kxz{t-1} \Kzz^{-1},\\
     \label{eq:Bt}
\mbB_{t-1} & := \Kxx{t-1} - \Kxz{t-1} \Kzz^{-1} \Kzx{t-1}, 
\end{align}
 % ----------------
 % $\mbA_{t-1}=\Kxy{\mbx_{t-1}}{\mbz}\Kxx{\mbz}^{-1}, \mbB_{t-1} = \Kxx{\mbx_{t-1}}-\Kxy{\mbx_{t-1}}{\mbz}\Kxx{\mbz}^{-1}\Kxy{\mbz}{\mbx_{t-1}}$.
and the cross-covariance term $\Kxz{t-1}:= \kernel(\mbx_{t-1}, \mbZ; \gphyper)$ and similarly for $\Kzx{t-1}$. Finally, $\trace(\cdot)$ is the trace operator and, as defined at the beginning of \cref{sec:gpssm}, $\mbQ$ is the transition noise covariance. 

\subsubsection{Optimal Variational Posterior}
With this, we obtain the form of the optimal variational distribution $q^*(\mbu, \mbx_{0:T})$ up to a normalizing constant $Z_q$ as:
\begin{multline} \label{eq:optimal_ux}
\log q^*(\mbu, \mbx_{0:T})=\log p(\mbu \g \mbZ) + \log p(\mbx_0) \\
+ \sum_{t=1}^T \Big[ \log  p(\mby_t \g \mbx_t)  
  + 
  \log \Normal (\mbx_t; \mbmu_{x_t}, \mbQ) \\
  - \frac{1}{2} \trace(\mbQ^{-1}\mbB_{t-1})\Big] 
  + \log Z_q.
\end{multline}
Here we note that the function values $\mbf_{1:T}$ have, effectively, been marginalized variationally. 
The optimal joint posterior over inducing variables and state trajectories depends on the prior over the inducing variables $p(\mbu \g \mbZ)$ stemming from the \gls{GP} functional prior, the prior over the initial state $p(\mbx_0)$ and the conditional likelihood terms  $p(\mby_t \g \mbx_t)$. 
It also depends on the resulting transitions  mapping $\mbx_{t-1}$ to $\mbx_{t}$ via the densities $\Normal (\mbx_t; \mbmu_{x_t}, \mbQ)$, where $\mbmu_{x_t}$ depends on $\mbx_{t-1}$ in a nonlinear way, as specified by \cref{eq:At}. 
The final trace term, $\trace(\mbQ^{-1}\mbB_{t-1})$, can be seen as a regularization term acting on state transitions, encouraging higher transition variances and, therefore, helping prevent overfitting. 
%
% In Eq.~(\ref{eq:optimal_ux}), $p(\mbu), p(\mbx_0), \mathcal{N}(\mby_t\g\mbx_t)$ are the prior distributions for the corresponding variables, and $\mathcal{N}(\mbx_t;\mbA_{t-1}\mbu,\mbQ)$ is $\mbx_t$'s Gaussian state transition density, in the sparse-GPSSMs setting. 
% The trace term $-\text{Tr}(\pmb{Q}^{-1}\mbB_{t-1})$ regularizes the variance associated with each latent state $\mbx_t$. Maximizing $-\text{Tr}(\pmb{Q}^{-1}\mbB_{t-1})$ would increase latent states' variance $\mbQ$, which tends to give a vague prior for latent states and also prevent the Gaussian transition density $\mathcal{N}(\mbx_t;\mbA_{t-1}\mbu,\mbQ)$ from overfitting, and would shrink the variance of prediction function values at the same time, which is a desired property in making predictions. 

\subsubsection{Alternative Perspective} 
An alternative way to obtain the optimal joint posterior over state trajectories and inducing variable is by bounding the true log joint marginal 
$\log p(\mby_{1:T}, \mbx_{0:T}, \mbu \g \mbZ)$ using Jensen's inequality. We give the details in \cref{app:free-form-bound}. This  has been used by previous work in standard regression settings \citep[see, e.g.,][and references therein]{pmlr-v130-rossi21a}. 
However, our setting considers the more complex case of \glspl{GPSSM}.
Additionally, our development shows much more clearly the optimal nature of the variational posterior, as we have obtained it via calculus of variations.

% An alternative way to understand our variational distribution $q^*(\mbu,\mbx)$ is that we may use the Jensen's inequality to obtain a lower bound for the ground-truth posterior distribution $p(\mbu, \mbx_{0:T}\g\mby_{1:T})$:
% \begin{align} \label{eq:alternative-view}
% &\log p(\mbu, \mbx_{0:T}\g\mby_{1:T})=\log\int p(\mbu, \mbx_{0:T}, \mbf_{1:T}\g\mby_{1:T})\diff\mbf_{1:T}\nonumber \\
% & =\log\int p(\mbf_{1:T}\g\mbu,\mbx_{0:T}) p(\mbx_{0:T}, \mbu, |\mby_{1:T}, \mbf_{1:T})\diff\mbf_{1:T}\nonumber \\
% & \ge\int p(\mbf_{1:T}\g\mbu,\mbx_{0:T})\log p(\mbx_{0:T}, \mbu|\mby_{1:T}, \mbf_{1:T})\diff\mbf_{1:T} \nonumber \\
% & =\log q^*(\mbu, \mbx_{0:T})
% \end{align}
% \cite{NIPS2015_6b180037,pmlr-v130-rossi21a} have obtained similar results, however, their results are for sparse Gaussian process regression only, which assumes $\mbx_{0:T}$ is independent. Our approach in GPSSMs needs to deal with the complex dependencies in $\mbx_{0:T}$.

% \noindent\textbf{Free-choice of likelihood} Our \gls{FFVD} method is versatile and can be applied to a wide range of likelihoods, not just continuous Gaussian ones. As demonstrated in Eq.~(\ref{eq:optimal_ux}), our method models each observation individually, rather than being influenced by the specific format of the likelihood. This allows for greater flexibility and adaptability in its application.

\subsection{Posterior Sampling}
% We use stochastic gradient Hamiltonian Monte Carlo (SG-HMC)~\cite{chen2014stochastic,havasi2018inference}, which is a variant of Markov chain Monte Carlo~(MCMC) methods, to obtain posterior samples from the variational distribution $q^*(\mbu,\mbx_{0:T})$. SG-HMC obtains samples from the target distribution with stochastic gradients and without evaluating the Metropolis ratio. 
Having the form of the optimal posterior in \cref{eq:optimal_ux}, we can then set the latent variables $\mbPsi \defeq \{\mbu, \mbx_{0:T}\}$ and have $\tilde{q}(\mbPsi) \propto q(\mbPsi) \approx p(\mbPsi \g \mby_{1:T})$. Thus, we can draw samples from our approximate posterior using \acrlong{SGHMC} \citep[\acrshort{SGHMC};][]{chen2014stochastic,havasi2018inference} and the energy function $U(\mbPsi) = - \log p(\mbPsi, \mby_{1:T}) = - \log  p(\mbPsi \g \mby_{1:T}) + \log Z_q \approx - \log  \tilde{q}(\mbPsi) + \log Z_q$. 
Using this procedure, samples from the target distribution can be obtained even with noisy gradients (e.g., with mini-batches) without requiring the evaluation of  Metropolis ratios. Importantly, other variables such as \gls{GP} hyper-parameters $\gphyper$ and inducing locations $\mbZ$  can be easily included in $\mbPsi$ using suitable priors and incorporating them in our objective in \cref{eq:optimal_ux}. 
\subsubsection{Computational Cost \& Prior Whitening}
\label{sec:computational-cost}
An interesting aspect of \gls{GPSSM} models is that, despite their apparent Markovian nature, sampling at time $T$ requires conditioning on all the previous $T-1$ points. 
This is due to the non-parametric coupled \gls{GP} prior over the transition function, making inference in the full  model $\cO(T^3)$ in time. 
Sparse variational inference approaches, such as those based on inducing variable approximations, still require expectations over entire trajectories  and, unlike standard supervised i.i.d settings, their time complexity is inherently dependent on $T$. 
Evaluation of the \gls{SGHMC} objective in \cref{eq:optimal_ux} for sampling in our \gls{FFVD} algorithm is $\cO(M^2 T)$. this is the same cost as that attained for \gls{ELBO} evaluation in \gls{VCDT}, which like our \gls{FFVD}, models dependencies between state trajectories and inducing variables.

% We may have a further comparison between our FFVD and the VCDT. 
Here we expand on the details of the computational cost of our approach when compared to \gls{VCDT}. Our \gls{SGHMC} objective in \cref{eq:optimal_ux} is very similar to the \gls{ELBO} used in \gls{VCDT}. The \gls{ELBO} in \gls{VCDT} requires expectations over $q(\mbu)$ and $q(\mbx_t,\mbx_{t-1}|\mbu)$. Given the factorization assumptions and the Gaussian constraints on these distributions, these expectations are estimated straightforwardly via Monte Carlo samples. The time complexity of evaluating \cref{eq:optimal_ux} or the \gls{ELBO} in \gls{VCDT} once (using one sample) is the same, i.e., $\mathcal{O}(M^3+M^2T)$.

The overall time complexity of both algorithms, \gls{FFVD} and \gls{VCDT}, depends on (i) the number of samples (noting, again, that \gls{VCDT} also requires samples from the approximate posterior to estimate the gradients of the \gls{ELBO}) and (ii) the number of iterations (either the length of the \gls{SGHMC} chain in \gls{FFVD} or the number of epochs for gradient-based optimization in \gls{VCDT}). As described in  \cref{sec:experiment-details}, our experimental setting followed closely that of the original \gls{VCDT} paper, which used $S=100$ samples for training and $S=10^5$ for predictions. Similarly, we used $S=100$ samples for \gls{FFVD}. Remarkably, the number of iterations in our experiments for \gls{FFVD} was $50,000$ while for \gls{VCDT} was $200,000$ to achieve convergence. Furthermore, our analysis in \cref{sec:additional-results} shows that, in fact, our \gls{FFVD} algorithm converges in less than 10,000 iterations.
 
As in previous work \citep[see, e.g.,][]{hensman-mcmc-2015}, we have observed that whitening the prior over the inducing variables improves the performance of our algorithm. See details of our whitening procedure and the resulting unnormalized log posterior in \cref{sec:whitening,sec:whitened-version}. 
\subsection{Smoothing and Predictive Distributions}
\label{sec:predictions}
We are interested in estimating the smoothing distribution $p(\mbx_{0:T} \g \mby_{1:T})$ and the predictive distribution $p(\mby_{T+1:T^\prime} \g \mby_{1:T})$ for $T^\prime \geq T$. 
At the end of our \gls{SGHMC} procedure, we have $S$ samples from our joint approximate posterior $q(\mbx_{0:T}, \mbu) \approx p(\mbx_{0:T} \g \mby_{1:T}) $, i.e., $\{ \sample{\mbx_{0:T}}{s}, \sample{\mbu}{s} \}_{s=1}^S$ and, therefore, the smoothing distribution (i.e., the marginal posterior over  $\mbx_{0:T}$) is readily available through this Monte Carlo approximation. 

We can also make one-step-ahead predictions using our posterior samples. In particular, using \cref{eq:marg-conditional,eq:transition-model} we have that:
\begin{multline}
    \label{eq:transition-x-given-u}
    p(\mbx_{t} \g \mbx_{t-1}, \mbu, \mbZ) = \Normal(\mbx_t; \mx{t-1} + \mbA_{t-1} (\mbu - \mz), \\ \mbB_{t-1} + \mbQ),
\end{multline}
$\forall t>T$. Thus, replacing the values of $\mbA_{t-1}, \mbB_{t-1}$ using \cref{eq:At,eq:Bt} we can make predictions for the next state using samples as 
\begin{multline}
    \label{eq:mut-sigmat-sample}
     %\sample{\mbx_{T}}{s} \g \sample{\mbx_{T-1}}{s}, \sample{\mbu}{s} 
     %& \sim \Normal( \sample{\mx{T-1}}{s} + \sample{\Kxz{T-1}}{s} \Kzz^{-1} (\sample{\mbu}{s} - \mz), \sample{\Kxx{T-1}}{s} -
     %\sample{\Kxy{T-1}{Z}}{s} \Kzz^{-1} \sample{\Kxy{Z}{T-1}}{s} + \mbQ), \\
    \sample{\mbx_{t}}{s} \g \sample{\mbx_{t-1}}{s}, \sample{\mbu}{s} 
      \sim \Normal(\mbx_{t}; \sample{\mbmu_t}{s}, \sample{\mbSigma_t}{s}) \quad \text{with} \\
     \sample{\mbmu_t}{s} = \sample{\mx{t-1}}{s} + \sample{\Kxz{t-1}}{s} \Kzz^{-1} (\sample{\mbu}{s} - \mz), \\
     \sample{\mbSigma_t}{s}  = \sample{\Kxx{t-1}}{s} -
     \sample{\Kxy{t-1}{Z}}{s} \Kzz^{-1} \sample{\Kxy{Z}{t-1}}{s} + \mbQ.
 \end{multline}
 For a general likelihood model, we can sample the (noisy) targets using 
 $
 %\begin{equation}
    \sample{\mby_t}{s} \g \sample{\mbx_t}{s}  \sim  p(\mby_t \g \sample{\mbx_t}{s}, \likehyper).
 %\end{equation}
 $
In the case of a Gaussian conditional  likelihood, as described in \cref{sec:gpssm}, we can see that given samples from the latent state,  the predictive distribution is a Gaussian
\begin{multline}
  % \sample{\mby_T}{s} \g  \sample{\mbx_T}{s} \sim \Normal(\mbC \sample{\mbx_T}{s} + \mbd, \mbR).
    \sample{\mby_t}{s} \g \sample{\mbx_{t-1}}{s}, \sample{\mbu}{s} 
      \sim   \Normal(\mby_{t};  \mbC  \sample{\mbmu_t}{s} + \mbd, \\ \mbC \sample{\mbSigma_t}{s}  \mbC^\top
     + \mbR).
\end{multline}
where the samples of $\{\mbx_T^{(s)}, \mbu^{(s)}\}$ are readily available after running \gls{SGHMC} on their joint space. 

\section{Collapsing Inducing Variables}
\label{sec:collapsed-vi}
So far we have described our method to obtain samples from the optimal variational posterior over the joint distribution of state trajectories and inducing variables using  \cref{eq:optimal_ux} and \gls{SGHMC}. 
In this section we show that we can, in fact, integrate out the inducing variables $\mbu$ from our  joint variational distribution and obtain the optimal marginal  distribution for latent states $\mbx_{0:T}$. 
We start by retaking \cref{eq:optimal_ux} and isolating the terms that depend on the inducing variables,
\begin{multline} 
    \label{eq:collapse_x_based_v}
    q^{*}({\mbx}_{0:T})
    =
    \int_{{\mbu}} q^*({\mbu}, {\mbx}_{0:T}) d{\mbu}   \\
    = p({\mbx}_0) \prod_{t=1}^T 
    \left[ p(\mby_t \g {\mbx}_t) \exp(-\frac{1}{2} 
    \trace(\mbQ^{-1}\mbB_{t-1}))\right]  \\
    \int_{{\mbu}} p(\mbu \g \mbZ) 
    \prod_{t=1}^T 
    \Normal({\mbx}_{t}; \mbmu_{x_t}, \mbQ) \diff \mbu ,
\end{multline}
where we note that $\mbmu_{x_t}$, as defined as in \cref{eq:muxt}, depends on the inducing variables $\mbu$. With this, we can complete the square and identify the terms in the integral as products of Gaussian distributions, whose normalization constant is a Gaussian. Therefore, we have 
\begin{multline}
% =&p({\mbx}_0)/\Normal(\sum_t{\mbx}_t^{\top}\mbQ^{-1}\mbF_{t-1}; \mathbf{0}, \mbI + \sum_t \mbF_{t-1}^{\top}\mbQ^{-1}\mbF_{t-1})\nonumber \\
%     &\prod_t\left[\log p(\mby_t|{\mbx}_t)\exp(-\frac{1}{2}\trace(\mbQ^{-1}\mbB_{t-1}))\Normal({\mbx}_t;\mathbf{0},\mbQ)\right]
q^{*}({\mbx}_{0:T}) = p({\mbx}_0) 
\prod_t\Big[ p(\mby_t|{\mbx}_t) \exp(-\frac{1}{2} \trace(\mbQ^{-1}\mbB_{t-1})) \\ \Normal({\mbx}_t;\mx{t-1},\mbQ)\Big]
/ \Normal(\tilde{\mbx}; \mathbf{0}, \tilde{\mbSigma}_x  ),
\end{multline}
where the Gaussian in the denominator is determined by
\begin{align}
\tilde{\mbx} & = \sum_{t=1}^T \tilde{\mbA}_{t-1}^\top \mbQ^{-1} (\mbx_{t} - \mbm_{t-1}), \\
\tilde{\mbSigma}_x &= \mbI + \sum_{t=1}^T \tilde{\mbA}_{t-1}^\top \mbQ^{-1} \tilde{\mbA}_{t-1}, \\
\tilde{\mbA}_{t-1} &= \Kxz{t-1} (\Lz^\top)^{-1},
%\pmb{\tilde{F}}_{t-1}=K_{{\mbx}_{t-1},Z}(\pmb{L}_{Z,Z}^{\top})^{-1}, K_{Z,Z}^{-1}=(\pmb{L}_{Z,Z}^{\top})^{-1}\pmb{L}_{Z,Z}^{-1}
\end{align} 
where $\Lz$ is the Cholesky decomposition of $\Kzz$, i.e., 
$\Kzz = \Lz \Lz^\top$. We note here that $\tilde{\mbx}$ is actually a projection of the cumulative uncorrelated inputs $\mbQ^{-1} (\mbx_{t} - \mbm_{t-1})$ via the projection matrix $\Lz^{-1} \Kzx{t-1}$ and, therefore, $\tilde{\mbx} \in \bbR^M$. 

Furthermore, in this collapsed version, the terms unrelated to $\mbu$ are kept the same as those in the original optimal joint distribution in \cref{eq:optimal_ux}. The individual Gaussian transitions for the latent states are now $\mathcal{N}(\mbx_t; \mbm_{t-1},\mbQ)$ since $\mbu$ has been integrated out and we recall that $\mbm_{t-1}=\mean(\mbx_{t-1})$.  
The additional term, which is the inverse of a multivariate Gaussian distribution, records the cumulative projected trajectory with the corresponding projected variances.  Maximizing $q^*(\mbx_{0:T})$ would tend to minimize this density function, which pushes the latent states away from $\mathbf{0}$ and also decreases the values of the $\mbQ$ regulated variance. 

Our algorithm in this collapsed version runs \gls{SGHMC} using as energy function $\mbPsi(\mbx_{0:T}) = - \log q(\mbx_{0:T})$ to obtain samples $\{\mbx_{0:T}^{(s)}\}$ from the marginal $q^*(\mbx_{0:T})$ and then uses the closed-form expression for the conditional (\cref{sec:closed-form-inducing}) to obtain $\{\mbu^{(s)}|\mbx_{0:T}^{(s)}\}$.
\subsection{Advantages of Collapsed Algorithm} 
\label{sec:collapsed-advantages}
Collapsing the inducing variables $\mbu$ will generally tend to improve the convergence of our  algorithm, as we are required to sample from a significantly lower number of latent variables. 
The computational cost is similar to that of the  uncollapsed algorithm since despite avoiding the computation of $\Kzz^{-1}$ in $p(\mbu \g \mbZ)$, we require a similar term, $\Lz^{-1}$, in the evaluation of the  projected Gaussian distributions.  

It is important to emphasize one particular difference in our approach with respect to the closely related \gls{VCDT} algorithm of \citet{overcome-mean-field-gp}. 
As described before,  \citet{overcome-mean-field-gp} also propose a coupled joint posterior between state trajectories and inducing variables. 
Their factorization is $q_{\acrshort{VCDT}}  = q(\mbu) q(\mbx_{0:T} \g \mbu)$, and they impose additional Gaussian constraints on these densities. Our implicit assumed factorization is  $q_{\acrshort{FFVD}}  = q(\mbx_{0:T}) q(\mbu \g \mbx_{0:T} )$, which allows us to obtain the optimal variational distribution without imposing any additional parametric constraints, integrate out the inducing variables analytically and get a Monte Carlo approximation to the optimal marginal $q(\mbx_{0:T})$  via samples. 
We also provide an expression for the conditional $q(\mbu \g \mbx_{0:T} )$ in closed-form. 
Details %on this and our derivation of the collapsed version 
can be found in  \cref{sec:closed-form-inducing}. 

% has the following benefits: (1), the algorithm should have better convergence behaviour as less variables ; (2), the ELBO might get better as $q^*(\mbx) $ would be closer to its groundtruth posterior distribution. At the same time, we keep the computational cost the same as that of using inducing variables, which is $\mathcal{O}(M^2N)$. 

\subsection{\Gls{PMCMC}}
It is clear that the latent states are constructed in a Markovian manner when the  transition function $f(\cdot)$ is given, as the value of the current latent state is dependent on the previous latent state's value. 
Therefore, we can use \gls{PMCMC} methods \citep{particlemcmc} to infer the posterior distribution of the Markov structured latent states $\mbx_{0:T}$. 
This Bayesian treatment might improve performance over  \gls{SGHMC}, as it incorporates the sequential nature of the problem into the sampling algorithm. The advantages of \gls{PMCMC} over standard sequential Monte Carlo approaches have been documented previously, see. e.g., \citet{particlemcmc}. 
Here we note that \citet{NIPS2013_2dffbc47} also proposed a \gls{PMCMC}  treatment for $\mbx_{0:T}$. However, their algorithm 
% is not scalable and -> not true :-( 
is very different from ours as it is based on the fully-independent conditional approximation \citep[see, e.g.,][]{quinonero2005unifying}. More details can be found in \cref{sec:extended-related-work}. 
\section{Related Work}
% Our \gls{FFVD} approach is closely related with previous methods in the GPSSM, i.e., Variational-GPSSMs~(VGPSSM)~\cite{vi-gpssm}, Probabilistic Recurrent state-space models~(PRSSM)~\cite{PRSSM}, Idetification of GPSSM~(IGPSSM)~\cite{eleftheriadis2017identification},  and Variational coupled dynamics and trajectories~(VCDT)~\cite{overcome-mean-field-gp}, all of which developed variational inference methods for scalable inference.
We have already described the main differences between our method and  closely-related approaches throughout the paper, e.g., in \cref{sec:intro,sec:main-methods,sec:collapsed-vi}. 
We refer the reader to \cref{sec:extended-related-work} for more details. 
Other works have considered the \gls{GPSSM} in partially observable unstable settings  \citep{pmlr-v120-curi20a}, combined variational inference with the Laplace approximation \citep{lindinger2022laplace} or used sample-based inference with a reduced rank approximation \citep{svensson16}. 
With regards to scalable \glspl{GP}, we note they have been 
the subject of much research effort in machine learning, with extensions to more general frameworks such as compositional models \citep[see, e.g.,][]{pmlr-v37-wilson15,salimbeni-2017,Haibin2019,pmlr-v70-cutajar17a,havasi2018inference,pmlr-v130-rossi21a,pmlr-v97-duncker19a,pmlr-v80-heinonen18a,pmlr-v180-hegde22a,auzina2022latent,bui2018efficient}. 
Our approach can be seen as a generalization of the fully-Bayesian supervised learning method proposed by \citet{pmlr-v130-rossi21a} to state-space models, where we have included the non-trivial component of \gls{GP}-transition dynamics and have  proposed a collapsed optimal variational distribution for state trajectories. 

Other models for \glspl{GP} on sequential data have been proposed, see, for example, \citet{frigola2015thesis} for an excellent overview.  
Of interest here is the state-space model view of \glspl{GP} that for time-series data with $d_x=1$ and Markovian covariance functions can provide exact inference in linear time $\cO(T)$ \citep{solin2016stochastic}. 
This has been extended to non-Gaussian likelihood models and made more efficient using several  computational primitives \citep{pmlr-v80-nickisch18a}.
%
% PRSSM takes the prior distribution of latent states as their posterior distributions, which results in inaccurate posterior inference. VCDT and IGPSSM both assume the posterior distribution of each latent state be a Gaussian distribution, which is incorrect as we can see from Eq.~(\ref{eq:collapse_x_based_v}) and the posterior distributions of latent states are much more complicated than simple Gaussian distribution. It should be noted that the Evidence Lower Bound~(ELBO) in all the above approaches is not presented in an analytical form, which results in the need to use time-consuming Monte Carlo methods for approximating it. In our \gls{FFVD} approach, we do not need to use Monte Carlo methods for approximating ELBO since we have already obtained the optimal format of $q(\mbu, \mbx_{0:T})$.
%
More recently, there has been some work on using the signature kernel \citep{toth2020bayesian,Salvi_2021} within \gls{GP} models. In particular,  \citet{lemercier2021siggpde} generalize variational orthogonal features \citep{burt2020variational,hensman2017variational} to the sequential case, constructing inducing variables associated with the signature kernel that yield a variational inference algorithm that does not require any matrix inversion.

\begin{figure}[t]
\centering
\includegraphics[width =0.225 \textwidth]{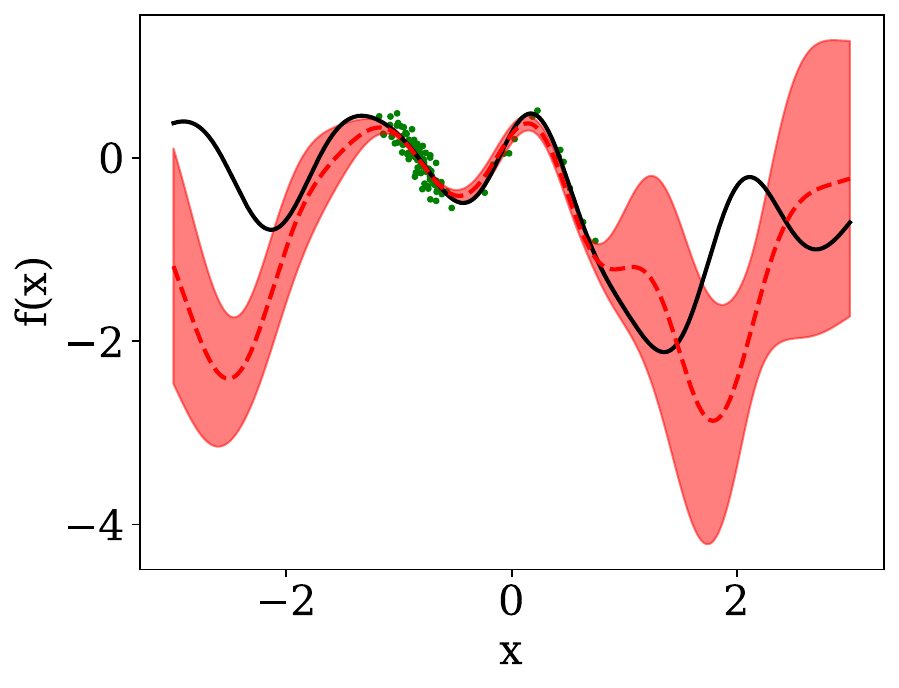}
\includegraphics[width =0.225 \textwidth]{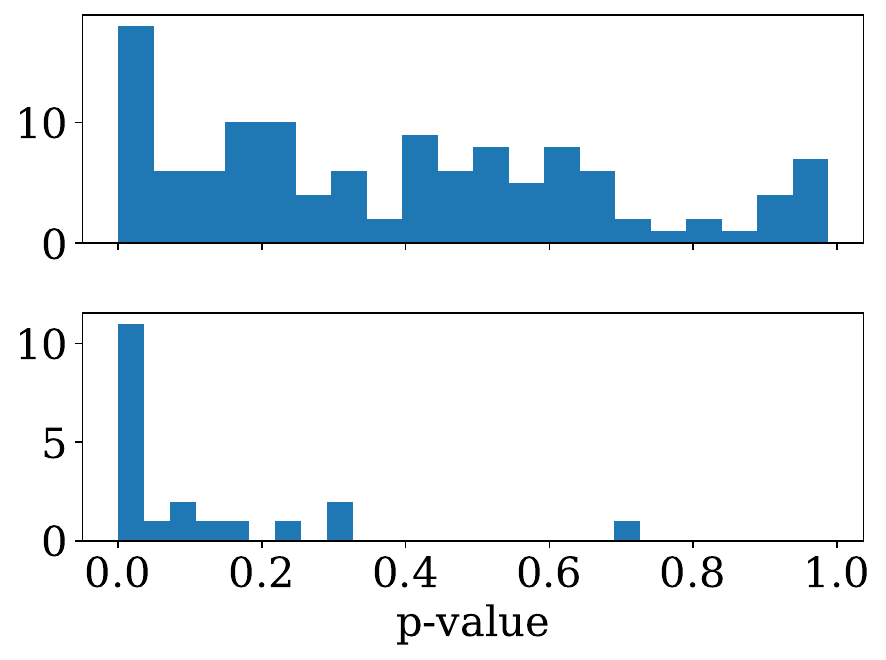}
\caption{Results on synthetic data. \emph{Left:} Observations shown as green dots,  Ground truth as a solid black line, and \gls{FFVD}'s mean fitting as a dashed red line with  one standard deviation error bars. 
\emph{Right:} Histograms of p-values for the hypothesis test that each  marginal posterior  over states $\{x_t\}$ (top) and inducing variables $\{u_i\}$ (bottom) is generated from a Normal distribution.}
\label{fig:results_visualisation}
\end{figure}

\begin{figure}[t]
\centering
\includegraphics[width =0.45 \textwidth]{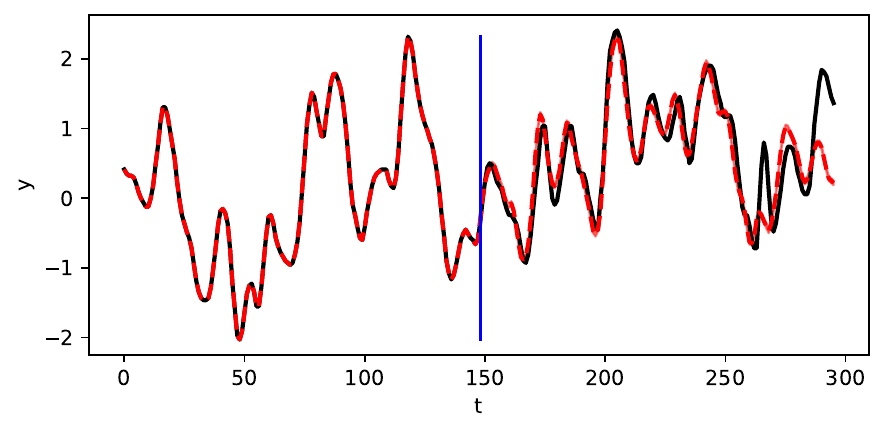}
\caption{Training ($t<=150$) and test performance ($t>=150$) on the Furnace dataset. The black solid line is the underlying ground truth signal and the dashed red line is \gls{FFVD}'s mean prediction. The blue solid line indicates the training/test split. An underlying $d_x=4$-dimensional latent stated was used.}
\label{fig:real-world-trajectory}
\end{figure}

\begin{figure}[t]
\centering
\includegraphics[width =0.4 \textwidth]{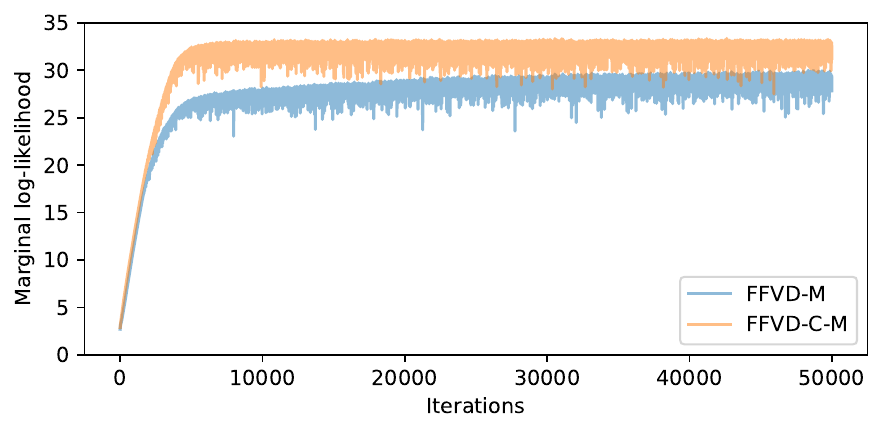}
\caption{Traceplot of the training log-likelihood when using \gls{SGHMC} (\acrshort{FFVDM}) and when collapsing the inducing variables  \acrshort{FFVDCM}.
Collapsing generally improved convergence.}
\label{fig:trace-plot-comparison-in-collapsed-version-one-data}
\end{figure}

%\eb{bring things commented out from the intro}
%\eb{say anything about deep net approaches to SSM?}

 % In fact, the variational parameter updating equations in VGPSSM~(Eq.~(10)(12) in~\cite{vi-gpssm}) share similar structures with Eq.~(\ref{eq:optimal_ux}) in here. The main difference is that VGPSSM used variational distributions to marginalized out corresponding random variables, either $\mbu$ or $\mbx_{1:T}$, which may result in an inaccurate approximation to the ground-truth posterior distribution.  
\section{Experiments}
\label{sec:experiments}
%To evaluate the effectiveness of our approach, we conduct experiments in two settings: using synthetic data to verify its properties, and using real-world data for fair comparisons with other methods.
We evaluate our \gls{FFVD} method on synthetic data and on six real-world system identification benchmarks \citep{overcome-mean-field-gp,PRSSM},   
comparing it with \acrshort{VGPSSM} \citep{vi-gpssm}, 
\acrshort{PRSSM} \citep{PRSSM}, 
\acrshort{VCDT} \citep{overcome-mean-field-gp}, 
and use a {LSTM} network \citep{hochreiter1997long}
as a baseline non-\gls{GP} based model. All experiment details can be found in \cref{sec:experiment-details}.

\begin{table*}[h!]
    \centering
    \caption{Test \acrfull{RMSE} values  $\pm$ one standard deviations on the real-world system identification benchmarks. Our method, \gls{FFVD}, when using \gls{SGHMC} (\gls{FFVDM}); the collapsed version (\gls{FFVDCM}); and when using \gls{PMCMC} (\gls{FFVDP}). } 
    \label{tab:rmse-value}
    \begin{tabular}{l c c c c c c}
    \toprule
  Methods & Actuator & Ballbeam & Drive & Dryer & Flutter & Furnace  \\
    \toprule
 \acrshort{LSTM} & $0.586\pm 0.411$  & $0.027\pm 0.023$   & $0.537\pm 0.108$    & $0.115\pm 0.029$ & $0.912\pm 0.562$   & $1.261\pm 0.610$   \\
\midrule
 \acrshort{VGPSSM} & $0.580 \pm 0.274$ & $0.073\pm 0.011$ & $0.722 \pm 0.087$ & $0.241 \pm 0.023$ & $1.482 \pm 0.218$ & $1.115 \pm 0.358$\\
 \acrshort{PRSSM} &  $0.497 \pm 0.381$ & $0.059 \pm 0.013$ & $0.813 \pm 0.101$ & $\mathbf{0.017} \pm 0.042$ & $1.371 \pm 0.156$ & $1.243 \pm 0.407$\\
\acrshort{VCDT} & $\mathbf{0.239}\pm 0.040$ & $0.011\pm 0.002$ & $0.585\pm 0.017$ & $0.142\pm 0.003$ & $1.782\pm 0.324$ & $1.166\pm 0.011$  
\\
 \midrule
 % \gls{FFVD} (optimized)  & $ 0.355\pm 0.103$ & $ 0.017\pm 0.013$ & $ 0.663\pm 0.259$ & $ 0.113\pm 0.021$ & $ 0.485\pm 0.168$ & $ 0.590\pm 0.204$        \\
 \acrshort{FFVDM} & $ 0.358\pm 0.242$ & $ 0.019\pm 0.018$ & $ 0.673\pm 0.207$ & $ 0.205\pm 0.313$ & $ \mathbf{0.280}\pm 0.193$ & $ 0.571\pm 0.185$    \\
%  \gls{FFVD} (sg-hmc (including z))  & $ 0.355\pm 0.228$ & $ 0.030\pm 0.059$ & $ 0.924\pm 4.499$ & $ 0.181\pm 0.157$ & $ 0.892\pm 5.136$ & $ 0.678\pm 0.371$  \\
% \gls{FFVD} (optimized + collapsed)   & $ 0.277\pm 2.668$ & $ 0.027\pm 0.012$ & $ \mathbf{0.638}\pm 0.099$ & $ 0.109\pm 1.832$ & $ 0.629\pm 0.217$ & $ 0.589\pm 0.248$   \\
\acrshort{FFVDCM}  & $ \underline{0.259}\pm 0.209$ & $ \mathbf{0.009}\pm 0.011$ & $ 0.775\pm 1.615$ & $ \underline{0.065}\pm 0.112$ & $ 0.663\pm 0.189$ & $ \mathbf{0.548}\pm 0.051$   \\
\acrshort{FFVDP}  & $ {0.388}\pm 0.087$ & ${0.199}\pm 0.045$ & $ \mathbf{0.342}\pm 0.057$ & $ {0.317}\pm 0.050$ & $ 0.562\pm 0.088$ & $ {0.669}\pm 0.174$   \\
 \bottomrule
      \end{tabular}
\end{table*}
\begin{table*}[h!]
    \centering
    \caption{Test \acrfull{NMLL} values $\pm$ one standard deviation on the real-world system identification benchmarks. Methods as in \cref{tab:rmse-value}.} 
    \label{tab:nlpd}
    \begin{tabular}{l c c c c c c}
    \toprule
  Methods & Actuator & Ballbeam & Drive & Dryer & Flutter & Furnace  \\
    \toprule
{\acrshort{VGPSSM}} & ${1.09} \pm 0.11$ & ${0.92} \pm 0.07$ & ${-0.60} \pm 0.07$ & $  0.46 \pm 0.09$ & $ 2.38 \pm 0.21$ & $  2.40 \pm 0.27$
\\  
{\acrshort{PRSSM}} & $0.29 \pm 0.17$ & $0.40 \pm 0.09$ & $\mathbf{-1.14} \pm 0.13$ & $  {0.52} \pm 0.07$ & $  0.51 \pm 0.11$ & $  3.46 \pm 0.31$
\\  
{\acrshort{VCDT}} & $\mathbf{-0.36} \pm 0.02$ & $\mathbf{-0.65} \pm 0.01$ & $1.23 \pm 0.01$ & $ { -0.02} \pm 0.01$ & $ { 6.13} \pm 0.48$ & $ { 7.49} \pm 0.07$
\\  
{\acrshort{FFVDM}} & $ { -0.03} \pm 0.10$ & $ { 0.09} \pm 0.05$ & $  {1.66} \pm 0.01$ & $ \mathbf{-0.08} \pm 0.09$ & $  \mathbf{0.48} \pm 0.33$ & $  \mathbf{-0.43} \pm 0.05$ \\  
 \bottomrule
      \end{tabular}
\end{table*}

\begin{table}[h!]
    \centering
    \caption{Mean training running times in seconds (T) and mean RMSE (R) as a function of the number of iterations on \textit{Furnace}.} \label{tab:running-time}
    \begin{tabular}{l c c c c c }
    \toprule
  Iterations & &$10$ & $100$ & $500$ & $1\,000$  \\
    \toprule
\multirow{2}{*}{{\acrshort{VGPSSM}}} & T &$\mathbf{0.71}$ &  $\mathbf{5.84}$ &  $\mathbf{27.96}$ &   $\mathbf{53.25}$\\  
& R & $1.58$& $1.49 $& $1.37$& $1.35$\\
 \midrule
\multirow{2}{*}{\acrshort{PRSSM}} & T & $17.44$ &  $205.91$ &  $1\,128.02$ &   $2\,137.59$\\
& R & $1.64$& $1.60$& $1.41$& $1.40$ \\
 \midrule
\multirow{2}{*}{\acrshort{VCDT}} & T & $26.96$ &  $269.71$ &  $1\,613.25$ &   $3\,374.78$\\  
& R & $\mathbf{1.39}$& $1.44$& $1.33$ & $1.25$\\
 \midrule
\multirow{2}{*}{\acrshort{FFVDM}}  & T & $3.19$ &  $26.33$ &  $121.42$  &  $238.68$\\  
& R & $1.50$ & $\mathbf{1.44}$ & $\mathbf{0.74}$ & $\mathbf{0.75}$\\
 \bottomrule
      \end{tabular}
\end{table}
\vspace{-0.2cm}
\subsection{Synthetic Data}
% We use the Gaussian process-generated random function as the transition function $f(\cdot)$ in the GPSSM.
We generate data from a sparse \gls{GPSSM} with a squared exponential covariance function. Our goal here is to investigate the properties of our \gls{FFVD} algorithm. Therefore,  we fix the value of all parameters to their ground-truth values except for the latent states $\mbx_{0:T}$ and the inducing variables $\mbu$. 
% Our algorithm estimates the posterior over these variables. 
%
The left panel of \cref{fig:results_visualisation} illustrates that  \gls{FFVD} effectively learns the intricate transition function with two modes within regions containing latent states. 
As the number of latent states increases,  \gls{FFVD} more accurately approximates the true function. 
A lack of fit in regions without latent states is to be expected. 
Having a good fit, we are now interested in knowing whether the true posterior (as estimated by our algorithm) is close to a Gaussian distribution. 
The right panel in \cref{fig:results_visualisation} indicates that more than $10\%$ of the latent states $\mbx_{0:T}$ and more than $50\%$ of the inducing variables $\mbu$ do not provide enough support for the hypothesis that their marginal distributions follow a Normal distribution (see \cref{sec:experiment-details}). This brings into question the parametric assumptions over the variational posterior made by previous work \citep[e.g.,][]{overcome-mean-field-gp, PRSSM}.
%
% We define three transition functions for the synthetic data case: (1) linear function, in which we define it as $f(x):=0.8x+\epsilon, \epsilon\sim\mathcal{N}(0, 0.1)$; (2) Kink function. This setting is the same as \cite{overcome-mean-field-gp}, and $f(\cdot)$ is defined as $f(x):=0.8 + (x+0.2)(1-\frac{5}{1+\exp(-2x)}), \epsilon\sim\mathcal{N}(0, 0.1)$, and (3) Gaussian process random function, in which we use the hyperparameters as. For all these cases
%We plot the random function approximation through the studies of VGPSSM, VCDT, PRSSM and our \gls{FFVD} methods. The detail visualisation is shown in Fig.~\ref{fig:function_fitting_performance}
% 

\subsection{Real-World Data}
Here we evaluate the different methods using six system identification benchmarks with a latent state dimension $d_x=4$, as used by \citet{overcome-mean-field-gp,PRSSM}. See more details in  \cref{sec:experiment-details}. 

\textbf{Predictive performance}: The  test performance is shown in \cref{tab:rmse-value} and \cref{tab:nlpd}, where we see that \acrshort{FFVD} attains the best \acrshort{NMLL} values in three out of the six benchmarks and obtains lowest 
\acrshort{RMSE} values in four out of the six benchmarks (in bold). 
Furthermore, for \emph{Actuator} and \emph{Dryer},  
\acrshort{FFVDCM} ranks second among all the algorithms (underlined). 
The performance of \acrshort{VGPSSM} and \acrshort{PRSSM} is usually worse than others, which is likely due to their strong mean-field and parametric assumptions, respectively. 
\acrshort{LSTM} obtains good performance on three datasets, although its deterministic structure is different from our random function setting. 
The performance of \acrshort{VCDT} is the closest to our \acrshort{FFVD} methods, as it models a coupled $q(\mbx_{0:T}, \mbu)$. 

\textbf{Qualitative analysis}: In addition to this quantitative evaluation, we  can see a qualitative illustration of using our algorithm for predicting the training and test (future) observations in \cref{fig:real-world-trajectory}. This is an example of good generalization, although it is (of course) not consistent across all problems, given the limited training data.
Finally, we analyze the convergence of our algorithm in  \cref{fig:trace-plot-comparison-in-collapsed-version-one-data}, where we see that  \gls{FFVDCM} uses less iterations ($\sim8\,000$ iterations) than \gls{FFVDM}  ($\sim40\,000$ iterations) to achieve similar performance. 
% That is, our collapsed algorithm generally improved convergence.
This confirms the benefits of collapsing the inducing variables and  sampling only on the lower-dimensional space of state trajectories. We can see these analyses for all benchmarks in \cref{sec:all-performance,sec:all-trace-plots}.

\textbf{Running times}: We illustrate the advantages of our approach when considering running time as a function of the number of iterations in \cref{tab:running-time}. We can clearly see that our approach is around $10$ times faster than \acrshort{PRSSM}/\acrshort{VCDT} and obtains the best \acrshort{RMSE} values after $100$ iterations. \acrshort{VGPSSM} runs fastest at the expense of higher prediction errors. These time results were done using a Macbook Pro 2021 with $16$GB in memory, M1 chip, and $8$ cores. It is noted that caution must be taken with interpreting these results, as they depend on implementation specifics, computer architectures along with other practical details.

\section{Conclusions, Limitations and Future Work}
\label{sec:discussion}
We have presented \acrshort{FFVD},  a new  variational inference algorithm for  \acrshortpl{GPSSM}. % based on  variational inference under the inducing-variable formalism. 
Unlike previous approaches, \gls{FFVD} does not make any independence or parametric assumptions on the joint variational posterior over state trajectories and inducing variables $q(\mbx_{0:T}, \mbu)$ and, instead, represents the posterior via samples from the ``optimal" variational distribution. 
% Furthermore, we have shown that our formulation allows us to collapse the inducing variables and carry out inference in the lower dimensional space of state trajectories. 
% Given samples from trajectories, we can make predictions using the optimal conditional $q(\mbu \g \mbx_{0:T})$, for which we have derived a closed-form expression. 
% 
However, as described in \cref{sec:var-family}, despite having a free-form posterior over state trajectories and inducing variables, our approach is still an approximation in that it assumes the conditional posterior over the latent functions to be the same as the conditional prior. This is a customary and necessary (albeit questionable) assumption in scalable variational methods for general \gls{GP} models. 

Our method also assumes independent \glspl{GP} over the state dimensions. However, one can expect the dynamics to be correlated across dimensions, bringing a need for the so-called multi-output/multi-task \glspl{GP}. This has been a fairly extensive area of research and 
%, with early approaches assuming decomposable covariances (i.e., Kronecker type). Because these direct approaches scale poorly on the number of tasks (i.e., dimensions in our case), 
we believe a weight-space view of such models could be suitable, where dimensions are correlated through a weight matrix. Approximate inference algorithms in this setting can be developed in $\mathcal{O}(d_x)$, where $d_x$ is the state dimensionality, or even independent of $d_x$ \citep[see, e.g.,][in the context of Cox process models]{aglietti2019efficient}. As we would like to scale up to very high-dimensional spaces, we leave this % direction of research 
to future work.

%\section{Conclusion}
% Future work will investigate scalable multi-output \glspl{GP} in order to deal with very high-dimensional state representations in an efficient way. 

% future work 
\bibliographystyle{icml2023}
\bibliography{references}

%%%%%%%%%%%%%%%%%%%%%%%%%%%%%%%%%%%%%%%%%%%%%%%%%%%%%%%%%%%%%%%%%%%%%%%%%%%%%%%
%%%%%%%%%%%%%%%%%%%%%%%%%%%%%%%%%%%%%%%%%%%%%%%%%%%%%%%%%%%%%%%%%%%%%%%%%%%%%%%
% APPENDIX
%%%%%%%%%%%%%%%%%%%%%%%%%%%%%%%%%%%%%%%%%%%%%%%%%%%%%%%%%%%%%%%%%%%%%%%%%%%%%%%
%%%%%%%%%%%%%%%%%%%%%%%%%%%%%%%%%%%%%%%%%%%%%%%%%%%%%%%%%%%%%%%%%%%%%%%%%%%%%%%
\newpage
\appendix
\onecolumn
\input{appendix.tex}
\end{document}

%% file: preamble.tex
\usepackage{scalefnt,letltxmacro}
\LetLtxMacro{\oldtextsc}{\textsc}
\renewcommand{\textsc}[1]{\oldtextsc{\scalefont{1.10}#1}}
\usepackage[usenames,dvipsnames]{xcolor} % EVB (use MidNightBlue color) 

\usepackage[utf8]{inputenc}
\usepackage{natbib}
\usepackage{graphicx}
\usepackage{amsmath}
\usepackage{amssymb}
\usepackage[acronym,nowarn]{glossaries}
\setacronymstyle{long-sc-short}
\glsdisablehyper
\makeglossaries

\usepackage[colorlinks,linktoc=all]{hyperref}
\usepackage[all]{hypcap}
\hypersetup{citecolor=MidnightBlue}
\hypersetup{linkcolor=MidnightBlue}
\hypersetup{urlcolor=MidnightBlue}

\usepackage[capitalize,nameinlink]{cleveref}
\crefname{section}{\S}{\S\S}
\Crefname{section}{\S}{\S\S}
\usepackage[textsize=tiny]{todonotes}

%% file: preamble_icml.tex
\usepackage{microtype}
\usepackage{graphicx}
\usepackage{subfigure}
\usepackage{booktabs} % for professional tables

\usepackage{hyperref}

\usepackage[accepted]{icml2023}
\usepackage{amsmath}
\usepackage{amssymb}
\usepackage{mathtools}
\usepackage{amsthm}
\usepackage{multirow}
\theoremstyle{plain}

\theoremstyle{definition}

\theoremstyle{remark}

\usepackage[textsize=tiny]{todonotes}

%% file: math.tex
\newcommand{\mx}[1]{\mbm_{#1}}
\newcommand{\mz}{\mbm_Z}
\newcommand{\Kxy}[2]{\mbK_{#1, #2}}
\newcommand{\Kxx}[1]{\mbK_{#1}}
\newcommand{\Kxz}[1]{\mbK_{#1, Z}}
\newcommand{\Kzx}[1]{\mbK_{Z, #1}}
\newcommand{\Kzz}{\mbK_{Z}}

\DeclareMathOperator{\trace}{Tr}
\newcommand{\Lz}{\mbL_Z}
\newcommand{\GP}{{\mathcal{GP}}}
\newcommand{\kernel}{\kappa_f}
\newcommand{\mean}{m_f} 

\newcommand{\gphyper}{\mbtheta}
\newcommand{\likehyper}{\mbphi}

\newcommand{\mathbold}[1]{\ensuremath{\boldsymbol{\mathbf{#1}}}}

\newcommand{\g}{\,|\,}
\renewcommand{\d}[1]{\ensuremath{\operatorname{d}\!{#1}}}
\newcommand{\nestedmathbold}[1]{{\mathbold{#1}}}

\newcommand{\mba}{\nestedmathbold{a}}

\newcommand{\mbd}{\nestedmathbold{d}}

\newcommand{\mbf}{\nestedmathbold{f}}
\newcommand{\mbg}{\nestedmathbold{g}}

\newcommand{\mbm}{\nestedmathbold{m}}

\newcommand{\mbu}{\nestedmathbold{u}}
\newcommand{\mbv}{\nestedmathbold{v}}

\newcommand{\mbx}{\nestedmathbold{x}}
\newcommand{\mby}{\nestedmathbold{y}}
\newcommand{\mbz}{\nestedmathbold{z}}

\newcommand{\mbA}{\nestedmathbold{A}}
\newcommand{\mbB}{\nestedmathbold{B}}
\newcommand{\mbC}{\nestedmathbold{C}}

\newcommand{\mbH}{\nestedmathbold{H}}
\newcommand{\mbI}{\nestedmathbold{I}}

\newcommand{\mbK}{\nestedmathbold{K}}
\newcommand{\mbL}{\nestedmathbold{L}}

\newcommand{\mbQ}{\nestedmathbold{Q}}
\newcommand{\mbR}{\nestedmathbold{R}}

\newcommand{\mbX}{\nestedmathbold{X}}

\newcommand{\mbZ}{\nestedmathbold{Z}}

\newcommand{\mbepsilon}{\nestedmathbold{\epsilon}}

\newcommand{\mbmu}{\nestedmathbold{\mu}}

\newcommand{\mbphi}{\nestedmathbold{\phi}}

\newcommand{\mbtheta}{\nestedmathbold{\theta}}

\newcommand{\mbPsi}{\nestedmathbold{\Psi}}
\newcommand{\mbSigma}{\nestedmathbold{\Sigma}}

\newcommand{\cL}{\mathcal{L}}
\newcommand{\cN}{\mathcal{N}}
\newcommand{\cO}{\mathcal{O}}

\newcommand{\Cov}{\bbC\text{ov}}
\newcommand{\E}{\mathbb{E}}

\newcommand{\bbR}{\mathbb{R}}
\newcommand{\bbC}{\mathbb{C}}
\newcommand{\bbV}{\mathbb{V}}

\newcommand{\bbE}{\mathbb{E}}

\newcommand{\Normal}{\cN}
\newcommand{\defeq}{:=}

\newcommand{\sample}[2]{#1^{(#2)}}

\newcommand{\diff}{\mathrm{d}}
\newcommand{\parderiv}[2]{\frac{\partial #1}{\partial #2}}

\usepackage{pifont}
\newcommand{\cmark}{\ding{51}}%
\newcommand{\xmark}{\ding{55}}%

%% file: acronyms.tex
\newacronym[plural=gp\textnormal{s}, firstplural=Gaussian processes]{GP}{gp}{Gaussian process}
\newacronym{HMC}{hmc}{Hamiltonian Monte Carlo}
\newacronym{SGHMC}{sghmc}{stochastic gradient Hamiltonian Monte Carlo}
\newacronym{PMCMC}{pmcmc}{particle Markov chain Monte Carlo}
\newacronym{MCMC}{mcmc}{Markov chain Monte Carlo}

\newacronym{RRANK}{rrank}{reduced rank}
\newacronym{SSM}{ssm}{state-space model}
\newacronym{GPSSM}{gpssm}{Gaussian process state-space model}
\newacronym{VCDT}{vcdt}{variationally coupled dynamic trajectories}
\newacronym{VGPSSM}{vgpssm}{variational Gaussian process state space model}
\newacronym{PRSSM}{prssm}{probabilistic recurrent state space model}
\newacronym{IGPSSM}{igpssm}{identifiable Gaussian process state space model}
\newacronym{FFVD}{ffvd}{free-form variational dynamics}
\newacronym{FFVDM}{ffvd-m}{\gls{FFVD} + \acrshort{MCMC} }
\newacronym{FFVDCM}{ffvd-c-m}{\gls{FFVD} + collapsed + \acrshort{MCMC}}
\newacronym{FFVDP}{ffvd-p}{\gls{FFVD} + \acrshort{PMCMC}}
\newacronym{FIC}{fic}{fully-independent conditional}
\newacronym{LSTM}{lstm}{Long short-term memory}

\newacronym{DL}{dl}{deep learning}
\newacronym{VI}{vi}{variational inference}
\newacronym{ELBO}{elbo}{evidence lower bound}
\newacronym{KL}{kl}{Kullback-Leibler}

\newacronym{RMSE}{rmse}{root mean square error}
\newacronym{NMLL}{nmll}{negative mean log likelihood}

%% file: appendix.tex
%\appendix
\section{Basic Results on Gaussian Distributions}
\subsection{Conditional Gaussian Distribution}
Assuming the joint Gaussian distribution $p(\mbx) = \Normal(\mbx; \mbmu, \mbSigma)$ over the random vector $\mbx$ such that:
\begin{equation}
\mbx = 
\begin{bmatrix}
\mbx_a \\
\mbx_b
\end{bmatrix}
, \quad 
\mbmu = 
\begin{bmatrix}
\mbmu_a \\
\mbmu_b
\end{bmatrix}
, \quad
\mbSigma = 
\begin{bmatrix}
\mbSigma_{aa} &  \mbSigma_{ab}\\
\mbSigma_{ba} & \mbSigma_{bb}, 
\end{bmatrix}
\end{equation}
where $\mbSigma_{ba} = \mbSigma_{ab}^T$, we have that the conditional distributions are given by:
\begin{equation}
p(\mbx_b \g \mbx_a ) = 
\Normal(\mbx_b; \mbmu_b + \mbSigma_{ba} \mbSigma_{aa}^{-1} (\mbx_a - \mbmu_a), 
\mbSigma_{bb} - \mbSigma_{ba} \mbSigma_{aa}^{-1}  \mbSigma_{ab}   ).
\label{eq:cond-gauss}
\end{equation}
\subsection{Expectation over log of Normal Distribution}
With an approximate marginal posterior $q(\mbf_\star)$ and a Normal distribution $p(\mby \g \mbf_\star)$ of the form:
\begin{align}
    q(\mbf_\star) &= \Normal(\mbf_\star; \mbmu_\star, \mbSigma_\star  ),\\
    p(\mby \g \mbf_\star) &= \Normal(\mby; \mbf_\star, \mbSigma_y), 
\end{align}
we can compute
\begin{align}
    \E_{q(\mbf_\star)} \log p(\mby \g \mbf_\star)  &= \log \Normal(\mby; \mbmu_\star, \mbSigma_y) 
    + \trace(\mbSigma_y^{-1} \mbSigma_\star).
\end{align}

\section{Posterior Marginal in Sparse GP Models}
In variational sparse GP models we  usually have the joint Gaussian model $p(\mbf_\star, \mbu)$ and a posterior distribution over $q(\mbu)$ with 
\begin{align}
    p(\mbf_\star) &= \Normal(\mx{\star}, \Kxx{\star}), \\
    p(\mbu) &= \Normal(\mbu; \mz, \Kzz), \\
    \Cov(\mbf_\star, \mbu) &= \Kxz{\star}, \\
    q(\mbu) &= \Normal(\mbu; \mbmu_\mbu, \mbSigma_\mbu), \\
    q(\mbf_\star, \mbu) & \defeq q(\mbu) p(\mbf_\star \g \mbu), 
\end{align}
and we wish to compute:
\begin{equation}
    q(\mbf_\star) = \int q(\mbf_\star, \mbu) \d{\mbu},
\end{equation}
where we have omitted conditional dependencies on $\mbx_\star$ and $\mbZ$ for simplicity in the notation. For example this is necessary to compute the expectations over the conditional log likelihood term during inferencce or to make predictions on a new test point. 
Using \cref{eq:cond-gauss} we obtain:
\begin{equation}
    p(\mbf_\star \g \mbu) = \Normal(\mbf_\star; \mx{\star} + \Kxz{\star} \Kzz^{-1} (\mbu - \mz),\Kxx{\star} - \Kxz{\star} \Kzz^{-1} \Kzx{\star} ). 
\end{equation}
In order to integrate out over $q(\mbu)$ we know that the result is a Gaussian with mean and covariances obtained from the following linear transformation of $\mbu$:
\begin{equation}
    \mbf_\star =  \mx{\star} +  \Kxz{\star} \Kzz^{-1} (\mbu - \mz) + \mbepsilon, \quad
    \mbepsilon \sim \Normal(\mathbf{0}, \Kxx{\star} - \Kxz{\star} \Kzz^{-1} \Kzx{\star} ).
\end{equation}
Thus, we obtain:
\begin{align}
    q(\mbf_\star) &= \Normal\left(
        \mbf_\star; 
        \mx{\star} +  \Kxz{\star} \Kzz^{-1} (\mbmu_\mbmu - \mz), 
        \Kxz{\star} \Kzz^{-1}  \mbSigma_\mbu \Kzz^{-1}  \Kzx{\star} + \Kxx{\star} - \Kxz{\star} \Kzz^{-1} \Kzx{\star} 
    \right) \\
      &= \Normal\left(
        \mbf_\star; 
        \mx{\star} +  \Kxz{\star} \Kzz^{-1} (\mbmu_\mbmu - \mz), 
        \Kxx{\star} + \Kxz{\star} \Kzz^{-1}  \left(\mbSigma_\mbu  - \Kzz \right)  \Kzz^{-1} \Kzx{\star}   
        \right).
\end{align}
\section{Graphical Model}
\begin{figure}[t]
\centering
\includegraphics[width =0.95 \textwidth]{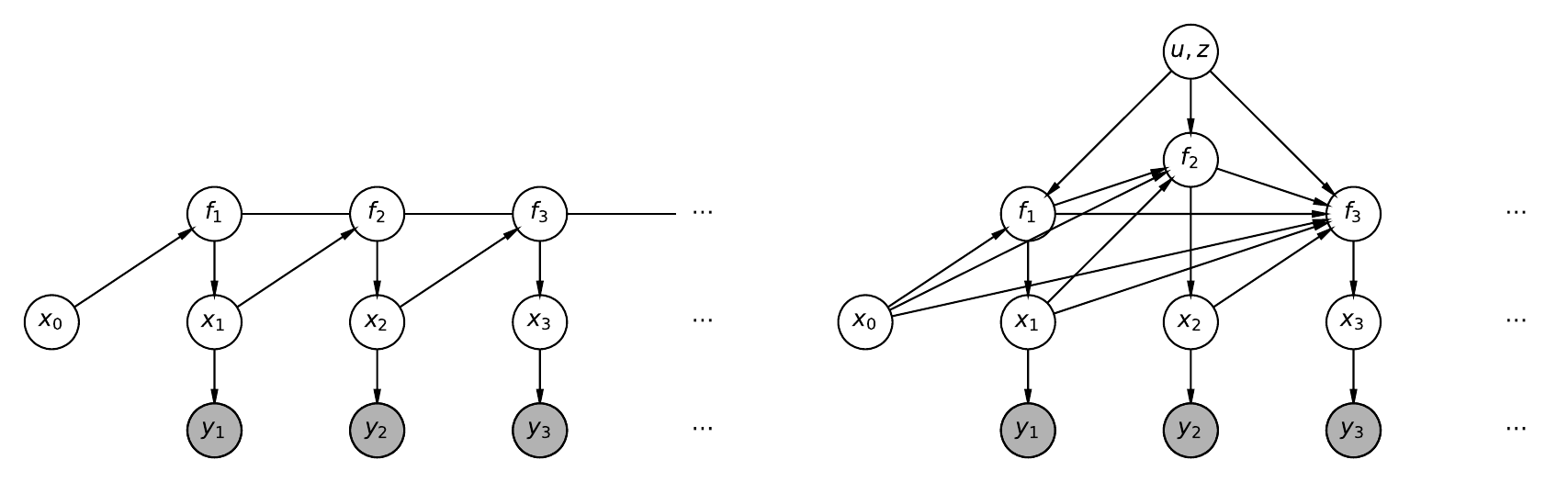}
\caption{Generative process of the \gls{GPSSM} (left panel) and the sparse \gls{GPSSM} (right panel).}
\label{fig:generative_process_GPSSM}
\end{figure}
\cref{fig:generative_process_GPSSM} illustrates \gls{GPSSM}'s graphical model.
\section{Details of Prior Reparameterization: Whitening Prior over Inducing Variables}
\label{sec:whitening}
\newcommand{\point}{\star}
Here we show a more detailed derivation of the prior reparameterization and, consequently, the new form of the required conditional distribution. We know that our prior over the inducing variables is $p(\mbu) = \Normal(\mbu; \mz, \Kzz)$ and that we use the new whitened prior $\mbv = \Lz^{-1} (\mbu - \mz)$ with $\Lz \Lz^T =\Kzz$ and $p(\mbv) = \Normal(\mbv; \mathbf{0}, \mbI_M)$. We also have that the marginal prior over $\mbf_\point$ is $p(\mbf_\point) = \Normal(\mbf_\point; \mx{\point}, \Kxx{\point})$. It follows that $p(\mbf, \mbv)$ is a Gaussian with cross-covariance:
\begin{align}
    \Cov(\mbf_\point, \mbv) &= \E [ (\mbf_\point - \mx{\point}) (\Lz^{-1}(\mbu -\mz) )^T ] \\
   &=  \underbrace{\E[ (\mbf_\point - \mx{\point}) (\mbu -\mz)^T ]}_{\Cov(\mbf_\point, \mbu)} (\Lz^{-1})^T \\
   &= \Kxz{\point} (\Lz^{-1})^T.
 \end{align}
Hence, using \cref{eq:cond-gauss}, we have that:
\begin{align}
    p(\mbf_\point \g, \mbx_\point, \mbv, \mbZ) &=
    \Normal(\mbf_\point; \mx{\point} + \Kxz{\point} (\Lz^{-1})^T \mbv, \Kxx{\point} - \Kxz{\point} (\Lz^{-1})^T \Lz^{-1} \Kzx{t} ), \\
    p(\mbf_\point \g, \mbx_\point, \mbv, \mbZ) &=
    \Normal(\mbf_\point; \mx{\point} + \Kxz{\point} (\Lz^{-1})^T \mbv, \Kxx{\point} - \Kxz{\point} \Kzz^{-1} \Kzx{\point} ).
\end{align}
Alternatively, we can simply obtain this result by replacing $\mbu$ with $\mbu = \mz + \Lz \mbv$ in the conditional distribution. Here we note that this is \emph{consistent} with the \gls{GP} definition in that if we were to integrate out the variables $\mbv$ from the joint model $p(\mbf_\point, \mbv)$ we would obtain the exact \gls{GP} prior $p(\mbf_\point)$.
\subsection{Posterior Marginal}
With the whitened join model $p(\mbf_\point, \mbv)$, Gaussian approximate posterior $q(\mbv)$ and a variationally sparse \gls{GP} model we have:
\begin{align}
    % p(\mbv) &= \Normal(\mbv; \mathbf{0}, \mbI_M), \\
    p(\mbf_\point \g \mbv) &=
    \Normal(\mbf_\point; \mx{\point} + \Kxz{\point} (\Lz^{-1})^T \mbv, \Kxx{\point} - \Kxz{\point} \Kzz^{-1} \Kzx{\point} ),\\
    q(\mbv) &= \Normal(\mbv; \mbmu_\mbv, \mbSigma_\mbv), \\
    q(\mbf_\star, \mbv) & \defeq q(\mbv) p(\mbf_\point \g \mbv).
\end{align}
It is easy to show that the posterior marginal $q(\mbf_\star) = \int q(\mbf_\star, \mbv)  \d{\mbv}$ is:
\begin{align}
    q_v(\mbf_\point) =\Normal(\mbf_\point; \mx{\point} + \Kxz{\point} (\Lz^{-1})^T \mbmu_\mbv , 
    \Kxx{\point} + \Kxz{\point} (\Lz^{-1})^T (\mbSigma_\mbv - \mbI_M) \Lz^{-1} \Kzx{\point}).
\end{align}
\section{Free-Form Posterior Estimation By Bounding the Log Marginal Likelihood Directly}
\label{app:free-form-bound}

Computing the log marginal:
\begin{equation}
        \log p(\mby_{0:T-1}, \mbx_{0:T-1}, \mbu \g \mbZ)
    = \log  \left[ p(\mbu \g \mbZ) 
    p(\mbx_0) p(\mby_0 \g \mbx_0 ) \prod_{t=1}^{T-1} p(\mby_t \g \mbx_t)  \right] + 
    \cL(\mbx_{0:T-1}, \mbu, \mbZ), 
\end{equation}
where 
\begin{equation}
\cL(\mbx_{0:T-1}, \mbu, \mbZ) \defeq
     \left[ \log \int_{\mbf_{1:T-1}} \prod_{t=1}^{T-1} p(\mbx_t \g \mbf_t) 
    p(\mbf_t \g \mbf_{1:t-1}, \mbx_{0:t-1}, \mbu, \mbZ) \ \mathrm{d}\mbf_{1:T-1} \right] ,
\end{equation}
which we note is a log of an expectation, which we can bounded using Jensen's inequality:
\begin{align}
    \cL(\mbx_{0:T-1}, \mbu, \mbZ)
    & \geq 
    \E_{\prod_{t=1}^{T-1} p(\mbf_t \g \mbf_{1:t-1}, \mbx_{0:t-1}, \mbu, \mbZ)  } 
    \log \prod_{t=1}^{T-1}  p(\mbx_t \g \mbf_t) \\
    & = \sum_{t=1}^{T-1} \E_{p(\mbf_t \g  \mbx_{t-1}, \mbu, \mbZ)}
        \log p(\mbx_t \g \mbf_t),
        \label{eq:e-log-xt}
\end{align}
where $p(\mbf_t \g  \mbx_{t-1}, \mbu, \mbZ)$ is the \gls{GP} predictive distribution over $\mbf_{t}$ at $ \mbx_{t-1}$ given the (pseudo observations) inducing variables $\mbu$ at inducing inputs $\mbZ$:
\begin{align}
    p(\mbf_t \g  \mbx_{t-1}, \mbu, \mbZ) 
    & = \Normal(\mbf_t; \mbA_{t-1} \mbu, \mbB_{t-1} ) \quad \text{with }\\
    \mbA_{t-1} &= \Kxz{t-1} \Kzz^{-1}  \\
    \mbB_{t-1} &= \Kxx{t-1} - \Kxz{t-1} \Kzz^{-1} \Kzx{t-1}.
\end{align}
% \begin{align}
%     \nonumber
%  \log p(\mby_{0:T-1}, \mbx_{0:T-1}, \mbu \g \mbZ) 
%  & \geq
%  \log  \left[ p(\mbu \g \mbZ) 
%     p(\mbx_0) p(\mby_0 \g \mbx_0 ) \prod_{t=1}^{T-1} p(\mby_t \g \mbx_t)  \right] + \\
% &  \qquad \int_{\mbf_{1:T-1}} \prod_{t=1}^{T-1}   p(\mbf_t \g \mbf_{1:t-1}, \mbx_{0:t-1}, \mbu, \mbZ)
%  \log   \prod_{t=1}^{T-1} p(\mbx_t \g \mbf_t) \ \mathrm{d} \mbf_{1:T-1} \\
% &=
% \nonumber
%  \log  \left[ p(\mbu \g \mbZ) 
%     p(\mbx_0) p(\mby_0 \g \mbx_0 ) \prod_{t=1}^{T-1} p(\mby_t \g \mbx_t)  \right] + \\
% & \qquad 
% \sum_{t=1}^{T-1} \E_{\prod_{\ell=1}^{T-1}   p(\mbf_\ell \g \mbf_{1:\ell-1}, \mbx_{0:\ell-1}, \mbu, \mbZ)} \log  p(\mbx_t \g \mbf_t)\\
% &=
% \nonumber
% \log  \left[ p(\mbu \g \mbZ) 
%     p(\mbx_0) p(\mby_0 \g \mbx_0 ) \prod_{t=1}^{T-1} p(\mby_t \g \mbx_t)  \right]
%     +
%     \sum_{t=1}^{T-1} \E_{p(\mbf_t \g  \mbx_{t-1})} p(\mby_t \g \mbx_t)
% \end{align}
Hence, the expectations in \cref{eq:e-log-xt} can be computed in closed-form:
\begin{equation}
   %\sum_{t=1}^{T-1} 
   \E_{p(\mbf_t \g  \mbx_{t-1}, \mbu, \mbZ)}
        \log p(\mbx_t \g \mbf_t)
        =
        \log \Normal(\mbx_t; \mbA_{t-1} \mbu, \mbQ) - 
        \frac{1}{2} \trace(\mbQ^{-1} \mbB_{t-1}).
\end{equation}
Thus, we have that:
\begin{multline}
    \log p(\mby_{0:T-1}, \mbx_{0:T-1}, \mbu \g \mbZ)
    \geq 
    \log  \left[ p(\mbu \g \mbZ) 
    p(\mbx_0) p(\mby_0 \g \mbx_0 )  \right]
    +\\
    \sum_{t=1}^{T-1} 
    \left[ 
        \log p(\mby_t \g \mbx_t) 
        + 
        \log \Normal(\mbx_t; \mbA_{t-1} \mbu, \mbQ) - 
        \frac{1}{2} \trace(\mbQ^{-1} \mbB_{t-1})
    \right].
\end{multline}
Then setting the latent variables $\mbPsi \defeq \{\mbu, \mbx_{0:T-1}\}$, we can obtain the approximate log unnormalized posterior:
\begin{multline}
    \log \tilde{q}(\mbPsi)
    \defeq \log  p(\mbu \g \mbZ) +
    \log p(\mbx_0) 
    + \log p(\mby_0 \g \mbx_0 ) 
    \ + \\
      \sum_{t=1}^{T-1} 
      \left[
      \log p(\mby_t \g \mbx_t) 
        + 
        \log \Normal(\mbx_t; \mbA_{t-1} \mbu, \mbQ)  -  
        \frac{1}{2} \trace(\mbQ^{-1} \mbB_{t-1})
            \right].
\end{multline}
where $\tilde{q}(\mbPsi) \propto q(\mbPsi) \approx p(\mbPsi \g \mby_{0:T-1})$. Thus, we can draw samples from our approximate posterior using \gls{SGHMC} and the energy function $U(\mbPsi) = - \log p(\mbPsi, \mby_{0:T-1}) = - \log  p(\mbPsi \g \mby_{0:T-1}) + C \approx - \log  \tilde{q}(\mbPsi) + C$. 
\subsection{Whitened Version}
\label{sec:whitened-version}
In the whitened version, $\mbv = \mbL_{Z}^{-1}(\mbu-\mbm_{Z})$ or $\mbu = \mbm_{Z}+\mbL_{Z}\mbv$ and the effective prior is 
\begin{equation}
    p(\mbv) = \Normal(\mbv; \mathbf{0}, \mbI). 
\end{equation}
Letting $\tilde{\mbu}=(\mbu, \mbZ, {\theta}), \tilde{\mbv}=(\mbv, \mbZ, {\theta})$, the determinant of the Jacobian can be calculated as:
\begin{align}
    |\frac{\partial \tilde{\mbu}}{\partial\tilde{\mbv}}|=&\left|\begin{array}{ccc}
        \frac{\partial (\mbm_{Z}+\mbL_{Z}\mbv)}{\partial \mbv} & \frac{\partial (\mbm_{Z}+\mbL_{Z}\mbv)}{\partial \mbZ} & \frac{\partial (\mbm_{Z}+\mbL_{Z}\mbv)}{\partial \theta} \\
        \frac{\partial \mbZ}{\partial \mbv} & \frac{\partial \mbZ}{\partial \mbZ} & \frac{\partial \mbZ}{\partial \theta} \\
        \frac{\partial \theta}{\partial \mbv} & \frac{\partial \theta}{\partial \mbZ} & \frac{\partial \theta}{\partial \theta} 
    \end{array}
    \right|\nonumber \\
    =&\left|\begin{array}{ccc}
        \mbL_{Z} & \frac{\partial (\mbm_{Z}+\mbL_{Z}\mbv)}{\partial \mbZ} & \frac{\partial (\mbm_{Z}+\mbL_{Z}\mbv)}{\partial \theta} \\
        0 & \mbI & 0 \\
        0 & 0 & \mbI 
    \end{array}
    \right|=\left|\begin{array}{ccc}
        \mbL_{Z} & 0 & 0 \\
        0 & \mbI & 0 \\
        0 & 0 & \mbI 
    \end{array}
    \right|=|\mbL_{Z}|=|\Kxx{Z}|^{\frac{1}{2}}
\end{align}
Thus, we can re-write the approximate log unnormalized posterior as:
\begin{multline}
    \log \tilde{q}(\mbv, \mbx_{0:T-1})=  \log |\Kxx{Z}|^{\frac{1}{2}}-\log |\Kxx{Z}|^{\frac{1}{2}}- \frac{1}{2} \mbv^\top \mbv +
    \log p(\mbx_0) 
    + \log p(\mby_0 \g \mbx_0 ) + \\
      \sum_{t=1}^{T-1} 
      \left[
      \log p(\mby_t \g \mbx_t) 
        + 
        \log \Normal(\mbx_t; \Kxz{t-1} (\Lz^{-1})^T \mbv, \mbQ)  -  
        \frac{1}{2} \trace(\mbQ^{-1} \mbB_{t-1})
            \right].\\
            = \log p(\mbv) +
    \log p(\mbx_0) 
    + \log p(\mby_0 \g \mbx_0 ) + \\
      \sum_{t=1}^{T-1} 
      \left[
      \log p(\mby_t \g \mbx_t) 
        + 
        \log \Normal(\mbx_t; \Kxz{t-1} (\Lz^{-1})^T \mbv, \mbQ)  -  
        \frac{1}{2} \trace(\mbQ^{-1} \mbB_{t-1})
            \right].
\end{multline}
%The same result can be obtained by optimizing the \gls{ELBO} wrt a free-form approximate posterior $q(\mbPsi)$ using calculus of variations, \todo{Add this.}
% where the posterior over all the latent variables is:
% \begin{equation}
%     q(\mbPsi, \mbf) = q(\mbx_{0:T-1}, \mbu, \mbf_{1:T-1}) \defeq 
%     q(\mbPsi) \prod_{t=1}^{T-1}  p(\mbf_t \g \mbf_{1:t-1}, \mbx_{0:t-1}, \mbu, \mbZ).
%\end{equation}

\section{Collapsed Method}
We can integrate out $\mbu$ in \cref{eq:optimal_ux} to obtain the marginal distribution of $\mbx_{1:T}$. 
\begin{align} 
\label{eq:marginal_x_step1}
    q^*(\mbx_{1:T})=
    \int_{\mbu} q^*(\mbu, \mbx_{1:T}) \diff \mbu 
    &= \int_{\mbu} p( \mbu \g \mbZ) p(\mbx_0) 
    \prod_{t=1}^T\left[ p(\mby_t \g \mbx_t)\exp(-\frac{1}{2}
    \trace(\mbQ^{-1}\mbB_{t-1})) 
    \Normal(\mbx_t; \mbmu_{x_t}, \mbQ)\right] \diff \mbu  \\
    &= p(\mbx_0)
    \prod_{t=1}^T \left[ p(\mby_t \g \mbx_t)
    \exp(-\frac{1}{2}\ \trace (\mbQ^{-1}\mbB_{t-1}))\right]
    \int_{\mbu} p( \mbu) 
    \prod_{t=1}^T \Normal 
    (\mbx_t; \mbmu_{x_t}, \mbQ) \diff \mbu. 
\end{align}
In the case of using our prior re-parameterization, i.e., whitened inducing variables, $\mbv$  and for multidimensional states and, therefore, multi-output \glspl{GP}, we can write the integral above for dimension $d$ as:
\begin{multline} 
\label{eq:whitened_v_multi-outputGP}
\int_{\mbv^{(d)}} p( \mbv^{(d)})
\prod_{t=1}^T
\Normal(\mbx_t^{(d)}; \mx{t-1}^{(d)} + \tilde{\mbA}^{(d)}_{t-1} \mbv^{(d)}, \mbQ_d) \diff \mbv^{(d)} \\
=  \prod_{t=1}^{T} \Normal 
(\mbx_t^{(d)};\mx{t-1}^{(d)},\mbQ_d)
\exp \left[\frac{1}{2}(\mbg^{(d)})^{\top}(\mbH^{(d)})^{-1}\mbg^{(d)}\right]\left(\det(\mbH^{(d)})\right)^{\frac{1}{2}}
\end{multline}
where 
\begin{align}
    \label{eq:g-mean}
    \mbg^{(d)} & := \left(\mbI+ \sum_{t=1}^{T} (\tilde{\mbA}_{t-1}^{(d)})^{\top} \mbQ_d^{-1}
    (\tilde{\mbA}_{t-1}^{(d)})\right)^{-1}
    \left(\sum_t(\mbx^{(d)}_t-\mx{t-1}^{(d)})^{\top}\mbQ_d^{-1} \tilde{\mbA}_{t-1}^{(d)} \right)^{\top},\\
    \label{eq:H-covariance}
    \mbH^{(d)} &= \left(\mbI+
    \sum_{t=1}^T (\tilde{\mbA}_{t-1}^{(d)})^{\top}\mbQ_d^{-1}(\tilde{\mbA}_{t-1}^{(d)})\right)^{-1}\nonumber \\
     \tilde{\mbA}_{t-1}^{(d)}  &= \Kxz{t-1}^{(d)} ((\Lz^{(d)})^{\top})^{-1}, \\
    \Kzz^{(d)} &= \Lz^{(d)} (\Lz^{(d)})^{\top}.
\end{align} 
Thus, our optimal log variational posterior marginal over state trajectories $\mbx_{0:T}$ is 
\begin{multline} 
\label{eq:collapse_x_based_v_app}
     \log q^{*}(\mbx_{0:T}) 
    = {\log p(\mbx_0)} 
     + \sum_{t=1}^{T} 
    \left[{\log p(\mby_t\g \mbx_t)}+ \log \Normal
    (\mbx_t;\mx{t-1},\mbQ)\right] \\
    -\frac{1}{2} \sum_{d=1}^{d_x} \left[ \sum_{t=1}^{T} 
    \trace(\mbQ_d^{-1} \mbB^{(d)}_{t-1})  
    % &- \frac{1}{2}\sum_d\left[\log\det\left(\mbI + 
    %  \sum_{t=1}^{T} (\tilde{\mbA}_{t-1}^{(d)})^{\top}\mbQ_d^{-1}(\tilde{\mbA}_{t-1}^{(d)})\right)\right.  \nonumber\\
    % & \left. + \frac{1}{2}\left(\sum_t(\mbx^{(d)}_t-\mx{t-1}^{(d)})^{\top}\mbQ_d^{-1}
    % \tilde{\mbA}_{t-1}^{(d)}\right)
    % \left(\mbI +
    % \sum_{t=1}^{T} (\tilde{\mbA}_{t-1}^{(d)})^{\top}\mbQ_d^{-1}(\tilde{\mbA}_{t-1}^{(d)})\right)^{-1}
    % \left(\sum_{t=1}^T (\mbx^{(d)}_t-\mx{t-1}^{(d)})^{\top}\mbQ_d^{-1} \tilde{\mbA}_{t-1}^{(d)}\right)^{\top}\right].
    +  \log \det (\mbH^{(d)})^{-1} 
    -   (\tilde{\mbx}^{(d)})^\top \mbH^{(d)}  (\tilde{\mbx}^{(d)}) \right],
\end{multline}
where
$
%\begin{equation}
\tilde{\mbx}^{(d)}  := \sum_{t=1}^T (\tilde{\mbA}_{t-1}^{(d)})^\top \mbQ_d^{-1} (\mbx_{t}^{(d)} - \mbm_{t-1}^{(d)}).
%\end{equation}
$
\section{Closed-Form Optimal Variational Conditional Distribution over Inducing Variables}
\label{sec:closed-form-inducing}
Given a trajectory, the optimal variational conditional over the whitened inducing variables given a state trajectory is given by:
\begin{equation}
    q(\mbv \g \mbx_{0:T})  = \prod_{d=1}^{d_x} \Normal(\mbv^{(d)}; \mbg^{(d)}, \mbH^{(d)}),
\end{equation}
where $\mbg$ and $\mbH$ are given in \cref{eq:g-mean,eq:H-covariance}, respectively. A similar expression can be obtained for the original (non-whitened) inducing variables $\mbu$. 
Of importance here is that our formulation has allowed us to obtain an optimal conditional variational distribution, given the state trajectories, over the inducing variables in closed form. 
We see then that, in effect, our approximate posterior is $q_{\acrshort{FFVD}}(\mbx_{0:T}, \mbu) = q(\mbx_{0:T}) q(\mbu \g \mbx_{0:T})$. 
In order to make predictions, we run \gls{SGHMC} to obtain samples $\{ \mbx_{0:T}^{(s)} \}$ from the marginal $p(\mbx_{0:T})$ and then use the closed-form expression for the conditional  to obtain $ \{\mbu^{(s)} \g \mbx_{0:T}^{(s)} \}$. Running \gls{SGHMC} over the much lower-dimensional space of state trajectories (instead of the joint space of trajectories and inducing variables) should generally converge faster. 
Or experiments with both versions of the algorithm confirm this. 
\subsection{Optimal Closed-Form Conditional vs Assumed Parametric Factorizations}
We contrast our variational distribution with that of the \acrlong{VGPSSM} \citep[\acrshort{VGPSSM};][]{vi-gpssm} and the \acrlong{VCDT} \citep[\acrshort{VCDT};][]{overcome-mean-field-gp}. 
While, implicitly, our variational distribution is $q_{\acrshort{FFVD}}(\mbx_{0:T}, \mbu) := q(\mbx_{0:T}) q(\mbu \g \mbx_{0:T})$ the \acrshort{VGPSSM} and \acrshort{VCDT} assume  $q_{\acrshort{VGPSSM}}(\mbx_{0:T}, \mbu) := q(\mbx_{0:T}) q(\mbu)$  and $q_{\acrshort{VCDT}}(\mbx_{0:T}, \mbu) := q(\mbx_{0:T} \g \mbu) q(\mbu)$, respectively.  
Although the former (\acrshort{VGPSSM}) obtained an ``optimal" posterior (given the factorization assumption), it is a mean-field approximation and ignores the dependencies between state trajectories and inducing variables.
The latter (\acrshort{VCDT}), does not make a mean-field assumption but its factorization forces parametric constraints over the individual distributions. 
In fact, \citet{overcome-mean-field-gp} assume Gaussian posteriors for both the marginal $q(\mbu)$ and each of the time-dependent conditionals in $q(\mbx_{0:T} \g \mbu)$. 
Our proposal is the only variational distribution that is theoretically optimal while yielding a closed-form conditional. 
\section{Particle \gls{MCMC}}
We can use the sequential structure of the latent states $\mbx_{0:T}$ for efficient inference. More specifically, using the transition in \cref{eq:transition-x-given-u}. The details of this are given in \cref{alg:PMCMC-for-X}.
\begin{algorithm}[t]
\caption{{Particle MCMC for inferring latent states $\mbx_{0:T}$}}
\label{alg:PMCMC-for-X}
\begin{algorithmic}
  \REQUIRE Number of particles $S$, observations $\mby_{1:T}$,  samples of $\mbx_{1:T}'$ in the previous iteration, likelihood parameters $ \mbphi = \{\mbC, \mbd, \mbR\}$, 
  % \ENSURE the samples of $\mbx_{1:T}^*$ for the current iteration
  \STATE Initialize state $\mbx_0$, weights $W^{(i)}_t=1$ for $i=1,\ldots, S$.
  \STATE Fix the last particle as setting $\mbx_{1:T}^{(S)} = \mbx_{1:T}'$
  \FOR{$t=1, \ldots, T$}
  \STATE Generate $\mbx_t^{(i)}$ from \cref{eq:transition-x-given-u} for $i=1, \ldots, S-1$.
  \STATE Calculate weight $W_t^{i}=p_{\mbphi}(\mby_t|\mbx_t^{(i)})$ for $i=1,\ldots, S$
  \STATE Normalize the weights as $\overline{W}_t^{(i)}=W_t^{(i)}/\sum_i W_t^{(i)}$
  \IF{$t<T$}
  \STATE For $i'=1, \ldots, S-1$, re-sample the index $j_{i'}$ from the categorical distribution, with the event probabilities being $(\overline{W}_t^{(1)}, \ldots, \overline{W}_t^{(S)})$. 
  \STATE For $i'=1, \ldots, S-1$, let $\mbx_{1:t}^{(i')}: =\mbx_{1:t}^{(j_{i'})}$ and set $W_t^{i'}=1/S$
  \ELSE 
  \STATE Re-sample the index $j^{*}$ from the categorical distribution, with the event probabilities being $(\overline{W}_t^{(1)}, \ldots, \overline{W}_t^{(S)})$.
  \STATE Return ${\mbx}_{0:T}:= =\mbx_{1:T}^{(j^*)}$
  \ENDIF
  \ENDFOR
\end{algorithmic}
\end{algorithm}
\section{Experiment Details}
\label{sec:experiment-details}
Regarding the \gls{VGPSSM}, \gls{PRSSM}, and \gls{VCDT}, we download the authors' implementations from their websites. 
For a fair comparison, we try to follow the same treatment for  hyper-parameters across all methods. We note, however, that the Matlab implementation of \gls{VGPSSM} does not provide inference for \gls{GP} hyper-parameters $\gphyper$, conditional likelihood parameters $\mbphi$, inducing inputs $\mbZ$, process variance $Q$ or observation variance $\mbR$.

\subsection{Settings for synthetic data}
In generating the synthetic data, we set the kernel's signal variance $\sigma=2.0$ and lengthscale $l=0.5$ in the sparse GPSSM. we reduce the impact of observational error and set the observation variance as small as $\sigma^2=0.01$, i.e., each observation would be generated as $y_t\sim\Normal(x_t,0.01)$. We also set the process variance $\mbQ=0.01$. 
The number of inducing points is set to $20$, and these $20$ inducing points are evenly spread in the interval $[-2, 2]$. We set the length of the training trajectory as $120$, the number of iterations as $50\,000$, and the number of posterior samples as $50$. 

\subsubsection{Testing the Marginals for Gaussianity} 
We conduct a hypothesis test for each individual trajectory state and inducing variable. Given the $50$ posterior samples for $\mbx_t$ (or $\mbu_m$), which we denote them as $\{\mbx_t^{(s)}\}_{s=1}^{50}$ (or $\{\mbu_m^{(s)}\}_{s=1}^{50}$), we use the implementation of \emph{scipy.stats.normaltest} from  Python's \emph{scipy} package to test whether we have sufficient evidence to reject the hypothesis that $\mbx_t$ (or $\mbu_m$) follows a Gaussian distribution. \emph{scipy.stats.normaltest} is based on the work of \citet{d1973tests}.  

\subsection{Settings for Real-world Data}
We adopt similar settings as in \citet{overcome-mean-field-gp} to initialize the hyper-parameters for all the models, which first uses a factorized nonlinear model to optimize $\gphyper, \mbx_{0:T}, \mbu, \mbZ, \mbC, \mbd, \mbR, \mbQ$. 
We use the identity function as the mean function and the squared exponential function with automatic relevance determination (ARD) in \gls{GPSSM}. 
When dealing with more than $1$ latent dimension, which requires a multi-output \gls{GP}, we use a different set of kernel hyper-parameters for each output. 
We set the number of inducing points to $M=100$ and the number of dimensions for latent states $\mbx_{0:T}$ to $d_x = 4$. 
 We set standard diagonal Gaussian priors for the initial latent state $\mbx_0$ and (where applicable) for likelihood parameters $\mbC, \mbd$, the logarithm of observation standard deviation $\log(\mbR)/2$ and process variance $\log(\mbQ)$. 
 
 For optimizing these hyper-parameters, we use Adam, with default settings for the optimizer parameters except for a decayed learning rate. We run our \gls{FFVD} algorithm for $50\,000$ iterations and \gls{VCDT} for $200\,000$. We set the learning rate as $0.01$ and the decay parameter as $0.05$ during \gls{SGHMC} sampling. We used $S=10^5$ posterior samples for \gls{VCDT} (as in the results in the original paper) and $S=100$ for ours.  
 Following the evaluation in \citet{overcome-mean-field-gp},  we use the \gls{RMSE}  between the models' predictions and the ground truth in the future $30$ time steps (not seen in training) as the comparison criterion for all the methods. 

Regarding the  benchmarks, we have $1\,024$ observations for \emph{Actuator}, $1\,000$ observations for \emph{Ballbeam}, $500$ observations for \emph{Drive}, $1\,000$ observations for \emph{Dryer}, $1\,024$ for \emph{Flutter}, and $296$ observations for \emph{Furnace}.
Training lengths are: $T_{\text{train}} = 500, 500, 250, 500, 500, 150$, respectively, and test lengths are the rest of the sequences. 
All these observations are $1$-dimensional but, as in previous work, we consider $d_x=4$-dimensional latent states. Each benchmark also contains a $1$-dimensional control input vector $\mba\in\mathbb{R}^{T\times 1}$. 

\subsection{Details of Performance Metrics}
We use the test \gls{RMSE} and the \gls{NMLL}. 

Given samples from the predictive distribution  $\{\hat{\mby}_t^{(s)}\}$ we can compute the \gls{RMSE} as:
\begin{equation}
    \gls{RMSE} = \frac{1}{T}  \sum_{t=1}^T 
     \left( \mby_t - \bbE\{\hat{\mby}_t^{(s)}\} \right)^2,
\end{equation}
where $\bbE\{\hat{\mby}_t^{(s)}\}$ is estimated as the empirical mean. 

If the corresponding method gives us $\{\mbx_t^{(s)}\}$, then we can estimate the $\gls{NMLL}$ as:
\begin{equation}
    \gls{NMLL} = - \frac{1}{S} \frac{1}{T} \sum_{s=1}^S \sum_{t=1}^T \log \Normal(\mby_t; \mbC \mbx_t^{(s)}  + \mbd, \mbR).
\end{equation}
If the method only provides us with $\{\hat{\mby}_t^{(s)}\}$ then a reasonable estimate is:
\begin{equation}
    \gls{NMLL} = - \frac{1}{T} \sum_{t=1}^{T} \log \Normal(\mby_t; \bbE\{\hat{\mby}_t^{(s)}\}, \bbV\{\hat{\mby}_t^{(s)}\}),
\end{equation}
where $\bbE\{\hat{\mby}_t^{(s)}\}, \bbV\{\hat{\mby}_t^{(s)}\}$ are the empirical mean and variance of the predicted samples. 

For \gls{FFVD}, we simply use:
\begin{equation}
    \gls{NMLL}_{\gls{FFVD}} = 
   - \frac{1}{S} \frac{1}{T} \sum_{s=1}^S \sum_{t=1}^T \log \Normal(\mby_{t};  \mbC  \sample{\mbmu_t}{s} + \mbd, \\ \mbC \sample{\mbSigma_t}{s}  \mbC^\top
     + \mbR),
\end{equation}
where $\sample{\mbmu_t}{s}$ and $\sample{\mbSigma_t}{s}$ are given by %\cref{eq:transition-x-given-u}. 
\cref{eq:mut-sigmat-sample}. 
\subsubsection{Data Normalization}
The normalization (i.e., standardization) of observations is given by the transformation:
\begin{equation}
    \tilde{y}_t = \frac{y_t -m_y}{\sigma_y},  
\end{equation}
where $m_y$ and $\sigma_y$ are the mean and standard deviation of the training data.  Let us denote the $\widetilde{\gls{RMSE}}$ and $\widetilde{\gls{NMLL}}$ as the error metrics in the normalized space. It is easy to show we can obtain the metrics in the original space as:
\begin{align}
    \gls{RMSE} &= \sigma_y \widetilde{\gls{RMSE}} \\
    \gls{NMLL} &= \log \sigma_y + \widetilde{\gls{NMLL}}.
\end{align}
\subsection{Reproducibility} 
%We will make our code publicly available upon acceptance of the paper.
As mentioned in the main paper, our code is publicly available at \url{https://github.com/xuhuifan/FFVD}.
\section{Additional Results}
\label{sec:additional-results}
% \subsection{Mean Negative Log Likelioods}
% \begin{table*}[h]
%     \centering
%     \caption{NMLL values for real-world datasets}
%     \label{tab:nmll-value}
%     \begin{tabular}{c|c|c|c|c|c|c}
%                 \hline
%   Methods & Actuator & Ballbeam & Drive & Dryer & flutter & furnace  \\
%         \hline
%  VCDT & $-0.3\pm 0.6$ & $-0.6\pm 0.9$ & $1.2\pm 0.9$ & $-0.1\pm 2.6$ & $7.4\pm 8.9$ & $7.5\pm 11.8$    \\
%  \hline
%   \acrshort{FFVDM} & $-0.0\pm 1.6$ & $16.8\pm 13.3$ & $2.0\pm 2.7$ & $1.0\pm 3.3$ & $28.3\pm 26.5$ & $14.4\pm 31.2$       \\
% \acrshort{FFVDCM} & $2982.8\pm 2483.9$ & $125.2\pm 263.4$ & $323.5\pm 919.6$ & $72.9\pm 87.9$ & $20863.6\pm 26524.1$ & $1907.3\pm 1776.9$  \\
% \acrshort{FFVDP} & $53.5\pm 58.5$ & $1757.2\pm 1849.4$ & $1345.4\pm 3343.8$ & $737.2\pm 866.5$ & $38744.9\pm 48415.9$ & $2357.2\pm 3540.1$   \\
%         \hline
%       \end{tabular}
% \end{table*}

\subsection{Training and Test Performance on All Benchmarks}
\label{sec:all-performance}
\begin{figure}[h]
\centering
\includegraphics[width =0.31 \textwidth]{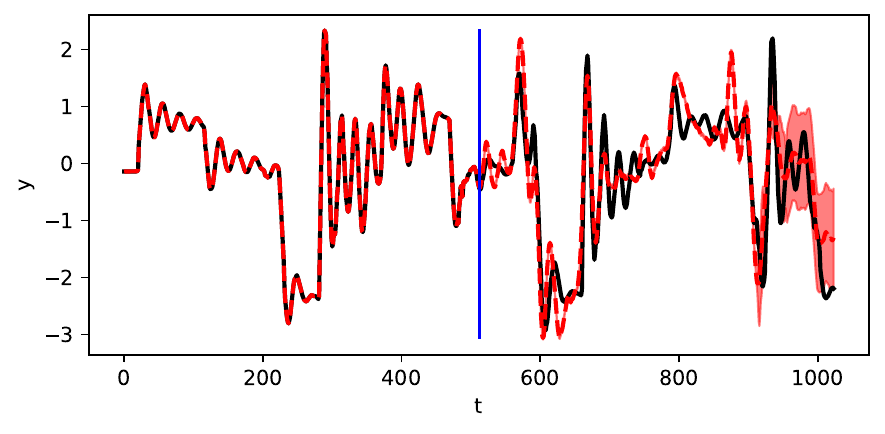}
\includegraphics[width =0.31 \textwidth]{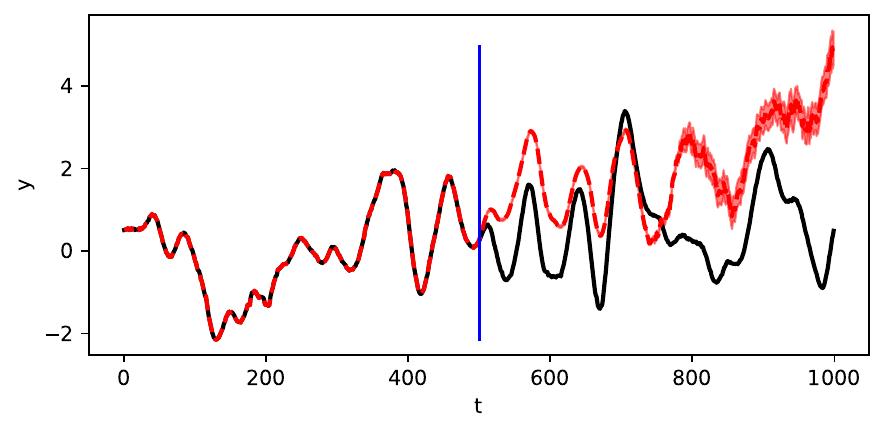}
\includegraphics[width =0.31 \textwidth]{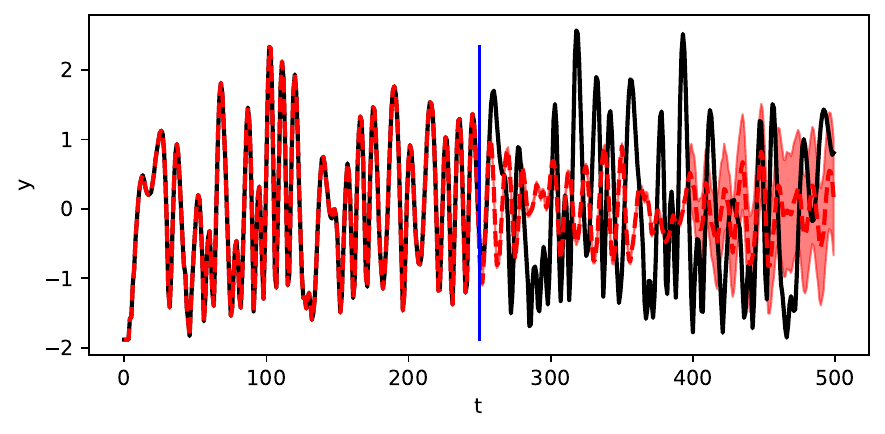}
\includegraphics[width =0.31 \textwidth]{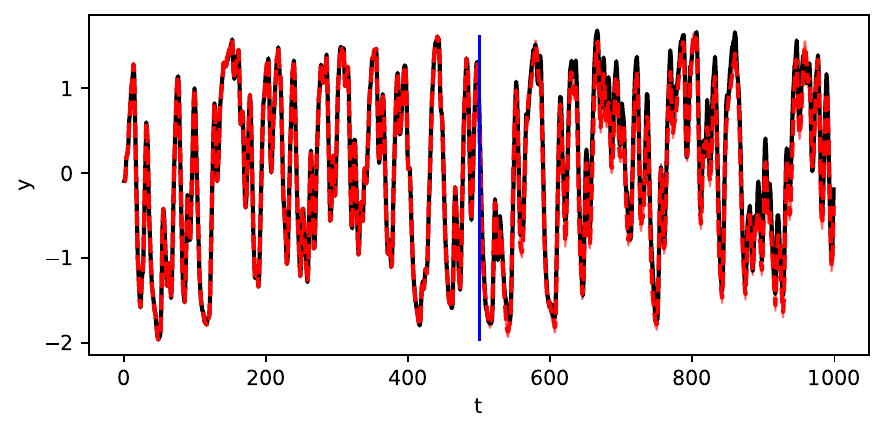}
\includegraphics[width =0.31 \textwidth]{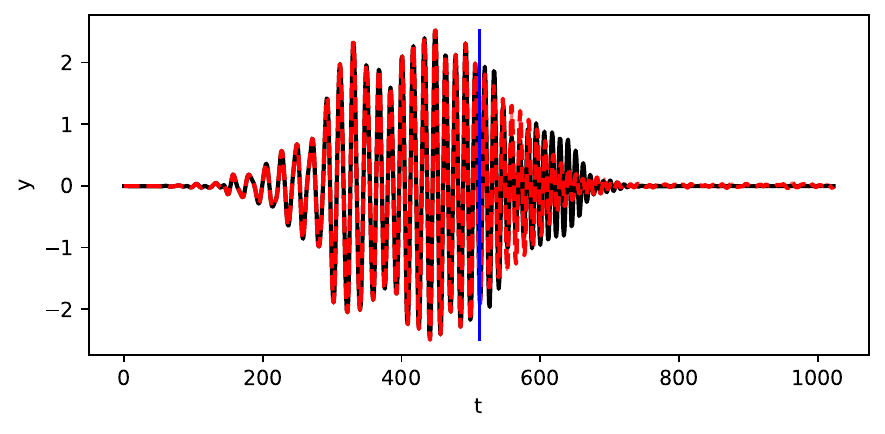}
\includegraphics[width =0.31 \textwidth]{individual_C2_VFE_Furnace_1_visualisation.pdf}
\caption{Training and test performance on benchmarks. Row 1 (left to right): \emph{Actuator}, \emph{Ballbeam}, \emph{Drive}; row 2 (left to right): \emph{Dryer}, \emph{Flutter}, \emph{Furnace}.}
\label{fig:trajectory-benchmarks}
\end{figure}

\cref{fig:trajectory-benchmarks} shows the training and testing regimes and predictions for the six system identification benchmarks considered. 

\subsection{Traceplot of Log-Likelihood on Training data for \gls{FFVDM} and \gls{FFVDCM}}
\label{sec:all-trace-plots}
\begin{figure*}[h]
\centering
\includegraphics[width =0.31 \textwidth]{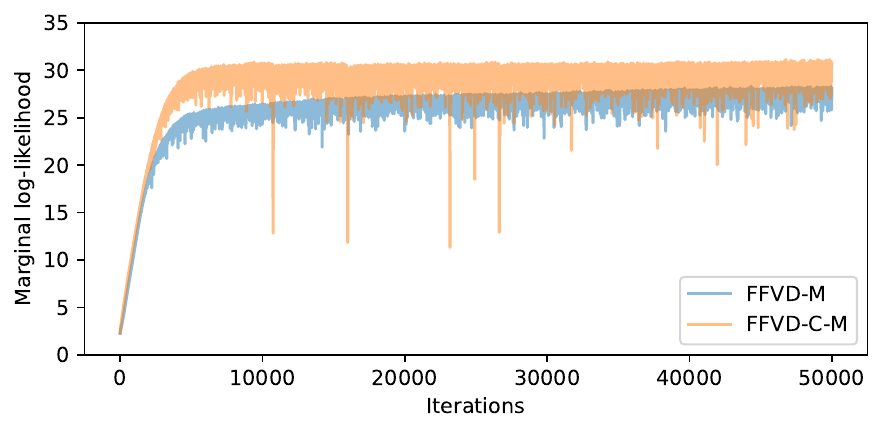}
\includegraphics[width =0.31 \textwidth]{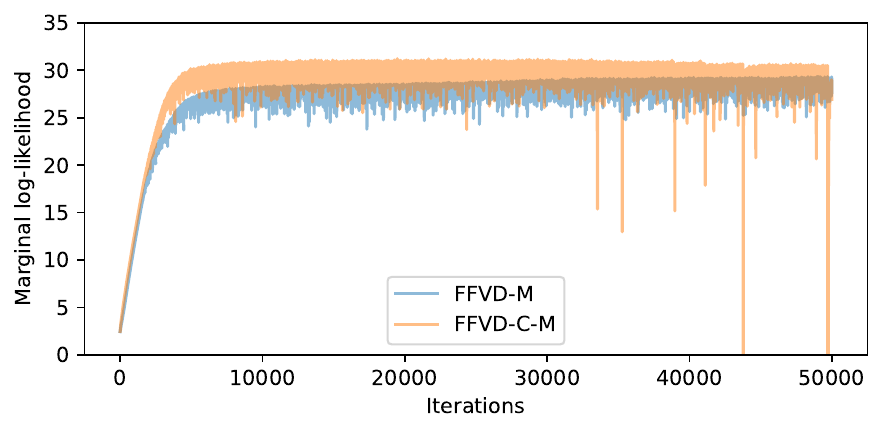}
\includegraphics[width =0.31 \textwidth]{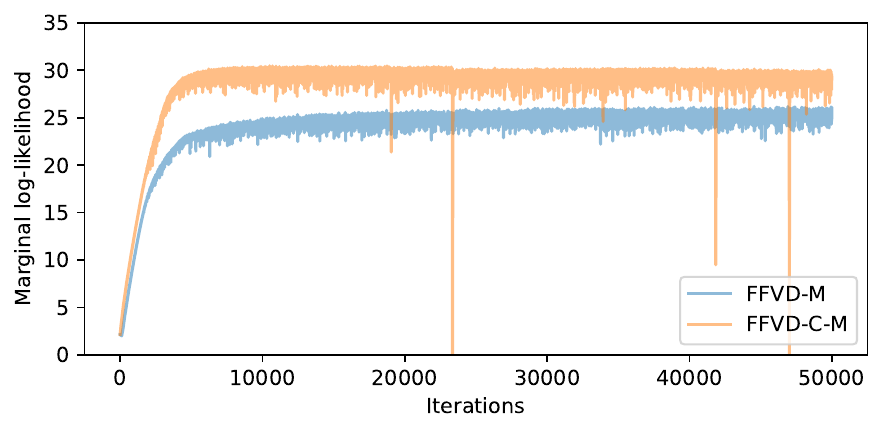}
\includegraphics[width =0.31 \textwidth]{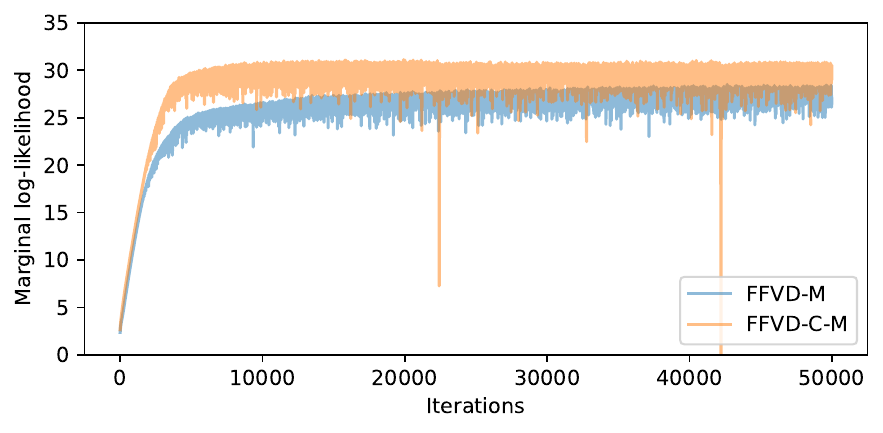}
\includegraphics[width =0.31 \textwidth]{sghmc_traceplot_VFE_Fluttervisualisation.pdf}
\includegraphics[width =0.31 \textwidth]{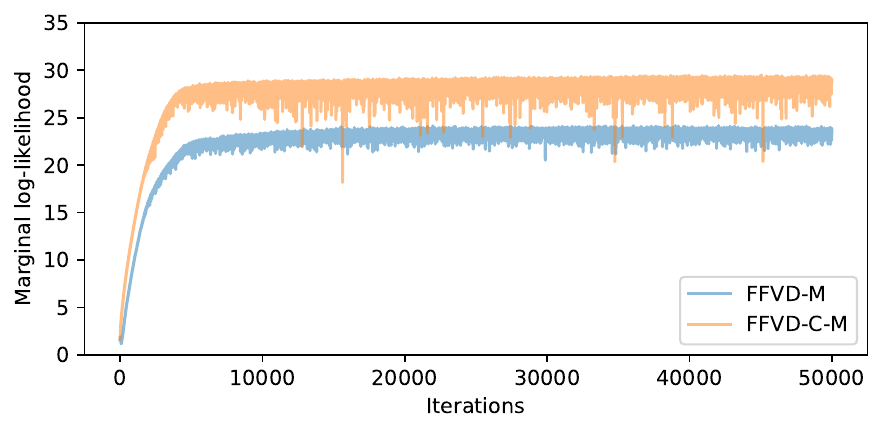}
\caption{Traceplot of log-likelihood on training data for FFVD-M and FFVD-C-M. Row 1 (left to right): \emph{Actuator}, \emph{Ballbeam}, \emph{Drive}; row 2 (left to right): \emph{Dryer}, \emph{Flutter}, \emph{Furnace}.}
\label{fig:trace-plot-comparison-in-collapsed-version}
\end{figure*}

\Cref{fig:trace-plot-comparison-in-collapsed-version} shows the convergence trace plots for all benchmarks. 

% \begin{figure*}[t]
% \centering
% \includegraphics[width =0.45 \textwidth]{aistatsimages/traceplot_VFE_Actuatorvisualisation.pdf}
% \includegraphics[width =0.45 \textwidth]{aistatsimages/traceplot_VFE_Ballbeamvisualisation.pdf}
% \includegraphics[width =0.45 \textwidth]{aistatsimages/traceplot_VFE_Drivevisualisation.pdf}
% \includegraphics[width =0.45 \textwidth]{aistatsimages/traceplot_VFE_Dryervisualisation.pdf}
% \includegraphics[width =0.45 \textwidth]{aistatsimages/traceplot_VFE_Fluttervisualisation.pdf}
% \includegraphics[width =0.45 \textwidth]{aistatsimages/traceplot_VFE_Furnacevisualisation.pdf}
% \caption{Test Log-likelihood for real world datasets}
% \label{fig:vfe}
% \end{figure*}

% \begin{table*}[t]
%     \centering
%     \caption{Experimental settings for VCDT, VFE}
%     \label{tab:experimental-settings}
%     \begin{tabular}{c|c|c}
%                 \hline
%    & VCDT & ourVFE   \\
%         \hline
%     Iterations & $2\times 10^5$ & $3\times 10^4$\\
%     Number of posterior samples & $10^5$ & $100$\\
%     Number of inducing points & $100$ & $100$ \\
%     Dimensions of latent states & $4$ & $4$ \\
%     Kernel type & RBF & RBF \\
%     Automatic-Relevance Determination & True & True \\
%     Prior for $\mbz$ &  & Normal \\
%     \hline
%       \end{tabular}
% \end{table*}

% \begin{table}[h]
%     \centering
%     \caption{Trainable status of random variables for different GPSSM methods}
%     \label{tab:traninable-variables}
%     \begin{tabular}{c|c|c|c|c|c|c|c|c}
%                 \hline
%   Methods & ${\mbu}$ & $Z$ & $X$ & $\phi$ & $Q$ & $C$ & $R$ \\
%         \hline
%  VGPSSM  & \cmark & {\xmark} & ${\mbx}_{0:T}$ & {\xmark} & {\xmark} & \cmark & {\xmark} \\
%  PRSSM  & \cmark & \cmark & ${\mbx}_0$ & \cmark & \cmark & \cmark & {\xmark} \\
%  VCDT  & \cmark & \cmark & ${\mbx}_{0:T}$ & \cmark & \cmark & \cmark & \cmark \\
%  \gls{FFVD}  & \cmark & \cmark & ${\mbx}_{0:T}$ & \cmark& \cmark & \cmark & \cmark \\
%         \hline
%     \end{tabular}
% \end{table}

%

\section{Extended Related Work}
\label{sec:extended-related-work}
Here we give more details about the differences between our method and closely related approaches for inference in the \gls{GPSSM}. 
The \gls{VGPSSM} of \citet{vi-gpssm} is the first developing variational  approaches to inference in \glspl{GPSSM}, and it uses mean-field assumptions to factorize the variational distribution of inducing variables $\mbu$ and latent state trajectories $\mbx_{0:T}$ as $q_{\gls{VGPSSM}}({\mbu}, \mbx_{0:T})=q({\mbu})q(\mbx_{0:T})$. 
This assumption clearly overlooks the complex dependencies between ${\mbu}$ and $\mbx_{0:T}$. \citet{PRSSM} model these dependencies with their \gls{PRSSM} algorithm, but make the unrealistic assumption of the posterior dynamics over the latent state trajectories being the same as the prior. \citet{overcome-mean-field-gp} show that this assumption can have critical consequences on the posterior and predictive distributions and propose a more general algorithm called \gls{VCDT}. Although a significant improvement over previous methods, \gls{VCDT} assumes posterior Gaussian distributions. When considering our main object of interest, i.e., the posterior over the state trajectories, under the standard and most commonly used \gls{GP} setting with non-linear kernels, a Gaussian posterior assumption never holds, as the latent states are inputs to the kernel. Hence, our approach is theoretically superior to \gls{VCDT} and provides a new baseline with broad practical applicability.

As mentioned in the main paper, \citet{NIPS2013_2dffbc47} also proposed a \gls{PMCMC} algorithm for inference over state trajectories. 
However, the algorithm is based on the \acrlong{FIC} approximation \citep[\acrshort{FIC};][]{quinonero2005unifying}. 
As reported in \citet{NIPS2013_2dffbc47}, a na\"{i}ve implementation of their algorithm has a time complexity of $\cO(M^2 T^2)$, which is significantly higher than ours. 
A much more efficient implementation with $\cO(M^2 T)$ time complexity that requires significant tracking of intermediate data structures and factorizations is hinted at in the original paper. 
Unfortunately, the corresponding code has not been made publicly available.  
Furthermore, model approximations such as those in \acrshort{FIC} can have surprising consequences, such as incorrectly estimating the noise variance to be almost zero or ignoring additional inducing inputs, see, e.g., \citet{bauer2016understanding} for a thorough discussion. Consequently, we see our variational approach as more principled. 

\subsection{Reduced-Rank Approaches}
Here we discuss the related approach of \citet{svensson16}, which we will refer to as \gls{RRANK}. As with random feature approximations, \gls{RRANK} exploits the covariance function-spectral density duality via the Wiener-Khinchin theorem to approximate the underlying \glspl{GP} with a linear-in-the-parameters model, where the feature maps can be obtained analytically for specific kernels.

While these types of approaches have their own weaknesses and strengths with respect to inducing variable approximations, one crucial difference with our method is that their approximation is only applicable to stationary kernels. Therefore our method is, by definition, much more general. Furthermore, as noted by \citet{Wilson2020}, those types of approximations to the covariance function based on finite-dimensional feature maps are known to exhibit undesirable pathologies at test time. In particular, the feature maps used by \citet{svensson16} can become increasingly ill-behaved in high-data regimes. Consequently, we believe the advantages of our approach (that avoids the stationarity assumption and its corresponding practical deficiencies) justify its use as an additional tool for practitioners and researchers.

Nevertheless, here we briefly report some results using the Matlab code provided by  \citet{svensson16}. The \gls{RMSE} values obtained on Actuator, Ballbeam, Drive, Dryer, Flutter, and Furnace are 0.33, 0.11, 0.73, 1.35, 1.86, and 9.66, respectively. While these results are generally competitive with our approach and other baselines, they are inferior to the results reported with our algorithm. 
% We obtained similar conclusions when using the NLPD metric.